\documentclass{article}

\usepackage[preprint]{corl_2022} 

\usepackage{url}            
\usepackage{booktabs}       
\usepackage{amsfonts}       
\usepackage{nicefrac}       
\usepackage{microtype}      
\usepackage{xcolor}         
\usepackage{amsmath}
\usepackage{graphicx, lipsum}
\usepackage{bbm}
\usepackage{wrapfig}
\usepackage{enumitem}
\usepackage{subcaption}
\usepackage{float}

\usepackage{algorithm}
\usepackage{algpseudocode}

\usepackage{tikz}
\newcommand{\cblock}[3]{
  \hspace{-1.5mm}
  \begin{tikzpicture}
    [
    node/.style={square, minimum size=10mm, thick, line width=0pt},
    ]
    \node[fill={rgb,255:red,#1;green,#2;blue,#3}] () [] {};
  \end{tikzpicture}%
}

\title{Watch and Match: Supercharging Imitation with Regularized Optimal Transport}

\makeatletter
\let\@fnsymbol\@arabic
\makeatother

\author{Siddhant Haldar\thanks{Correspondence to: sh6474@nyu.edu} \qquad Vaibhav Mathur \qquad Denis Yarats \qquad Lerrel Pinto \vspace{0.1in}
\\ \vspace{0.1in} New York University
\\{\small \tt \href{https://rot-robot.github.io/}{rot-robot.github.io}}
\vspace{-0.2in}
}

\begin{document}
\maketitle


\begin{abstract}
\label{abstract}

Imitation learning holds tremendous promise in learning policies efficiently for complex decision making problems. Current state-of-the-art algorithms often use inverse reinforcement learning (IRL), where given a set of expert demonstrations, an agent alternatively infers a reward function and the associated optimal policy. However, such IRL approaches often require substantial online interactions for complex control problems. In this work, we present Regularized Optimal Transport (ROT), a new imitation learning algorithm that builds on recent advances in optimal transport based trajectory-matching. Our key technical insight is that adaptively combining trajectory-matching rewards with behavior cloning can significantly accelerate imitation even with only a few demonstrations. Our experiments on 20 visual control tasks across the DeepMind Control Suite, the OpenAI Robotics Suite, and the Meta-World Benchmark demonstrate an average of $7.8\times$ faster imitation to reach $90\%$ of expert performance compared to prior state-of-the-art methods. On real-world robotic manipulation, with just one demonstration and an hour of online training, ROT achieves an average success rate of 90.1\% across 14 tasks.

\end{abstract}


\begin{figure}[h!]
\centering
\includegraphics[width=0.99\linewidth]{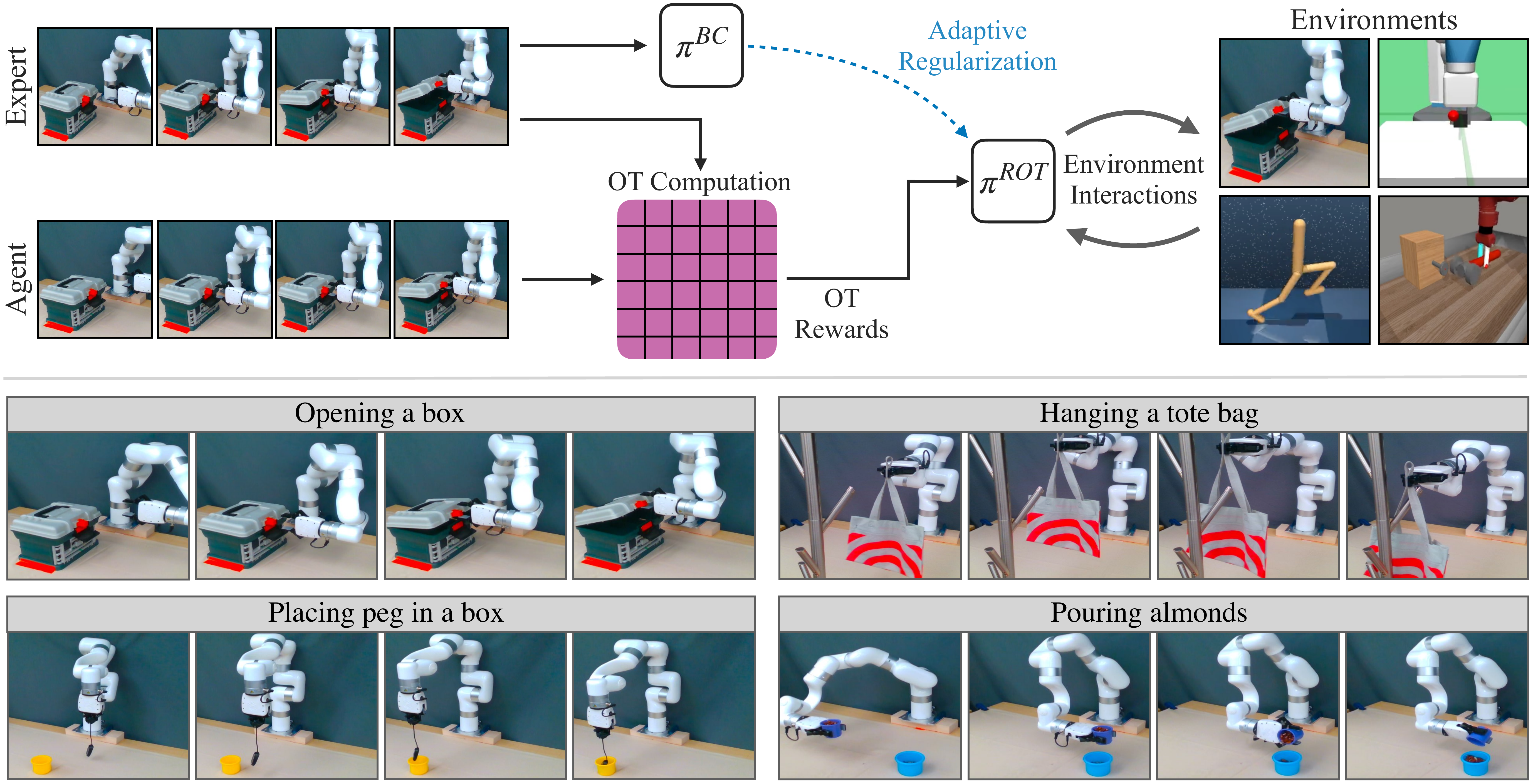}
\caption { \textbf{(Top)} Regularized Optimal Transport (ROT) is a new imitation learning algorithm that adaptively combines offline behavior cloning with online trajectory-matching based rewards. This enables significantly faster imitation across a variety of simulated and real robotics tasks, while being compatible with high-dimensional visual observation. \textbf{(Bottom)} On our xArm robot, ROT can learn visual policies with only a single human demonstration and under an hour of online training. }
\label{fig:intro}
\end{figure}

\section{Introduction}
\label{introduction}

Imitation Learning (IL)~\cite{pomerleau1998autonomous,andrychowicz2020learning,kolm2020modern} has a rich history that can be categorized across two broad paradigms, Behavior Cloning~(BC)~\cite{pomerleau1998autonomous} and Inverse Reinforcement Learning~(IRL)~\cite{ng2000algorithms}. BC uses supervised learning to obtain a policy that maximizes the likelihood of taking the demonstrated action given an observation in the demonstration. While this allows for training without online interactions, it suffers from distributional mismatch during online rollouts~\cite{ross2011reduction}. IRL, on the other hand, infers the underlying reward function from the demonstrated trajectories before employing RL to optimize a policy through online environment rollouts. This results in a policy that can robustly solve demonstrated tasks even in the absence of task-specific rewards~\cite{ho2016generative,kostrikov2018discriminator}.

Although powerful, IRL methods suffer from a significant drawback -- they require numerous expensive online interactions with the environment. There are three reasons for this: (a) the inferred reward function is often highly non-stationary, which compromises the learning of the associated behavior policy~\cite{kostrikov2018discriminator}; (b) even when the rewards are stationary, policy learning still requires effective exploration to maximize rewards~\cite{yarats2021mastering}; and (c) when strong priors such as pretraining with BC are applied to accelerate policy learning, ensuing updates to the policy cause a distribution shift that destabilizes training~\cite{nair2020awac,uchendu2022jump}. 
Combined, these issues manifest themselves on empirical benchmarks, where IRL methods have poor efficiency compared to vanilla RL methods on hard control tasks~\cite{cohen2022imitation}. 

In this work, we present Regularized Optimal Transport (ROT) for imitation learning, a new method that is conceptually simple, compatible with high-dimensional observations, and requires minimal additional hyperparameters compared to standard IRL approaches. In order to address the challenge of reward non-stationarity in IRL, ROT builds upon recent advances in using Optimal Transport~(OT)~\cite{papagiannis2020imitation,dadashi2020primal,cohen2022imitation} for reward computation that use non-parametric trajectory-matching functions. To alleviate the challenge of exploration, we pretrain the IRL behavior policy using BC on the expert demonstrations. This reduces the need for our imitation agent to explore from scratch. 

However, even with OT-based reward computation and pretrained policies, we only obtain marginal gains in empirical performance. The reason for this is that the high-variance of IRL policy gradients~\cite{schulman2017proximal,silver2014deterministic} often wipe away the progress made by the offline BC pretraining. This phenomenon has been observed in both online RL~\cite{rajeswaran2017learning} and offline RL~\cite{nair2020awac} methods. Inspired by solutions presented in these works, we stabilize the online learning process by regularizing the IRL policy to stay close to the pretrained BC policy. To enable this, we develop a new adaptive weighing scheme called soft Q-filtering that automatically sets the regularization -- prioritizing staying close to the BC policy in the beginning of training and prioritizing exploration later on. In contrast to prior policy regularization schemes~\cite{rajeswaran2017learning,jena2020augmenting}, soft Q-filtering does not require hand-specification of decay schedules. 

To demonstrate the effectiveness of ROT, we run extensive experiments on 20 simulated tasks across DM Control~\cite{tassa2018deepmind}, OpenAI Robotics~\cite{brockman2016openai}, and Meta-world~\cite{yu2020meta}, and 14 robotic manipulation tasks on an xArm (see Fig.~\ref{fig:intro}).
Our main findings are summarized below.
\begin{enumerate}[leftmargin=*,align=left]
    \item ROT outperforms prior state-of-the-art imitation methods, reaching $90\%$ of expert performance $7.8\times$ faster than our strongest baselines on simulated visual control benchmarks. 
    \item On real-world tasks, with a single human demonstration and an hour of training, ROT achieves an average success rate of 90.1\% with randomized robot initialization and image observations. This is significantly higher than behavior cloning (36.1\%) and adversarial IRL (14.6\%).
    \item ROT exceeds the performance of state-of-the-art RL trained with rewards, while coming close to methods that augment RL with demonstrations (Section~\ref{sec:reward_rl} \& Appendix~\ref{appendix:reward_rl}). Unlike standard RL methods, ROT does not require hand-specification of the reward function.
    \item Ablation studies demonstrate the importance of every component in ROT, particularly the role that soft Q-filtering plays in stabilizing training and the need for OT-based rewards during online learning (Section~\ref{sec:sqf-exp} \& Appendix~\ref{appendix:ablations}). 
\end{enumerate}

Open-source code and demonstration data will be publicly released on our project website. Videos of our trained policies can be seen here: \url{rot-robot.github.io/}.

\section{Background}
\label{background}
Before describing our method in detail, we provide a brief background to imitation learning with optimal transport, which serves as the backbone of our method. Formalism related to RL follows the convention in prior work~\cite{yarats2021mastering,cohen2022imitation} and is described in Appendix~\ref{appendix:background}. 

\paragraph{Imitation Learning with Optimal Transport (OT)} The goal of imitation learning is to learn a behavior policy $\pi^b$ given access to either the expert policy $\pi^e$ or trajectories derived from the expert policy $\mathcal{T}^e$. While there are a multitude of settings with differing levels of access to the expert~\cite{torabi2019recent}, our work operates in the setting where the agent only has access to observation-based trajectories, i.e. $\mathcal{T}^e \equiv \{(o_t, a_t)_{t=1}^{T}\}_{n=1}^N$. Here $N$ and $T$ denotes the number of trajectory rollouts and episode timesteps respectively. Inverse Reinforcement Learning (IRL)~\cite{ng2000algorithms,abbeel2004apprenticeship} tackles the IL problem by inferring the reward function $r^e$ based on expert trajectories $\mathcal{T}^e$. Then given the inferred reward $r^e$, policy optimization is used to derive the behavior policy $\pi^b$. 

To compute $r^e$, a new line of OT-based approaches for IL~\cite{papagiannis2020imitation,dadashi2020primal,cohen2022imitation} have been proposed. Intuitively, the closeness between expert trajectories $\mathcal{T}^e$ and behavior trajectories $\mathcal{T}^b$ can be computed by measuring the optimal transport of probability mass from $\mathcal{T}^b \rightarrow \mathcal{T}^e$. 
Thus, given a cost matrix $C_{t,t^{'}} = c(o^{\boldsymbol{b}}_{t},o^{\boldsymbol{e}}_{t'})$  and the optimal alignment $\mu^{*}$ between a behavior trajectory $o^{\boldsymbol{b}}$ and and expert trajectory $o^{\boldsymbol{e}}$, a reward signal for each observation can be computed using the equation:
\begin{equation}
    \label{eq:ot_reward}
    r^{OT}(o^{\boldsymbol{b}}_{t}) = - \sum_{t'=1}^{T} C_{t,t^{'}} \mu^{*}_{t,t^{'}}
\end{equation}
A detailed account of the OT formulation has been provided in Appendix~\ref{appendix:background}.

\paragraph{Actor-Critic based reward maximization} Given rewards obtained through OT computation, efficient maximization of the reward can be achieved through off-policy learning~\cite{kostrikov2018discriminator}. In this work, we use Deep Deterministic Policy Gradient (DDPG)~\cite{lillicrap2015continuous} as our base RL optimizer which is an actor-critic algorithm that concurrently learns a deterministic policy $\pi_\phi$ and a Q-function $Q_\theta$. 
However, instead of minimizing a one step Bellman residual in vanilla DDPG, we use the recent n-step version of DDPG from \citet{yarats2021mastering} that achieves high performance on visual control problems.

\section{Challenges in Online Finetuning from a Pretrained Policy}
\label{challenge}

\begin{figure}[t!]
\begin{center}
\includegraphics[width=\textwidth]{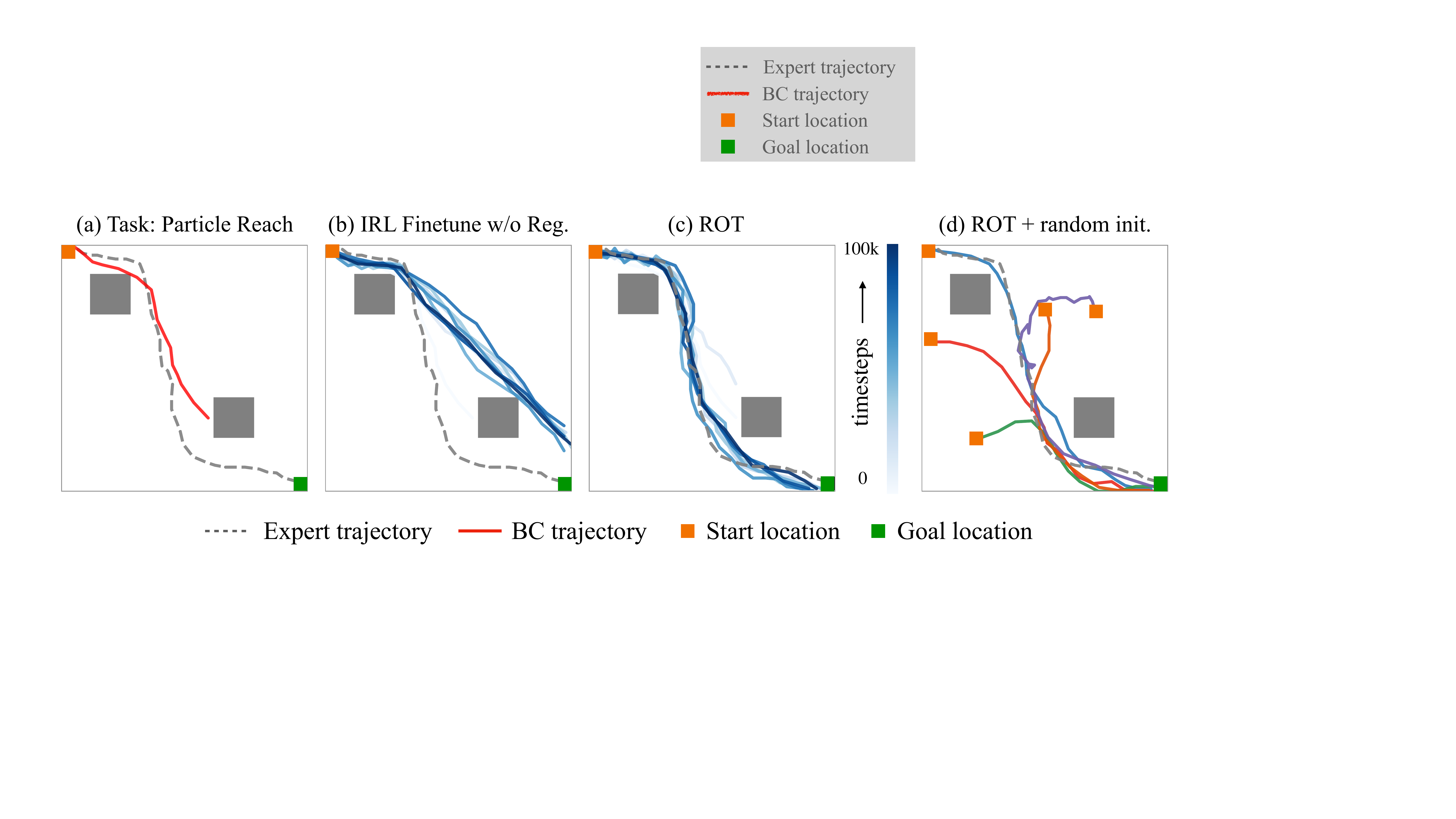} 
\caption{(a) Given a single demonstration to avoid the grey obstacle and reach the goal location, BC is unable to solve the task. (b) Finetuning from this BC policy with OT-based reward also fails to solve the task. (c) ROT, with adaptive regularization of OT-based IRL with BC successfully solves the task. (d) Even when the ROT agent is initialized randomly, it is able to solve the task.}
\label{fig:toy}
\end{center}
\end{figure}

In this section, we study the challenges with finetuning a pretrained policy with online interactions in the environment. Fig.~\ref{fig:toy} illustrates a task where an agent is supposed to navigate the environment from the top left to the bottom right, while dodging obstacles in between. The agent has access to a single expert demonstration, which is used to learn a BC policy for the task. Fig.~\ref{fig:toy}~(a) shows that this BC policy, though close to the expert demonstration, performs suboptimally due to accumulating errors on out-of-distribution states during online rollouts~\cite{ross2011reduction}. Further, Fig.~\ref{fig:toy}~(b) uses this BC policy as an initialization and naively finetunes it with OT rewards (described in Section~\ref{background}). Such naive finetuning of a pretrained policy (or actor) with an untrained critic in an actor-critic framework exhibits a forgetting behavior in the actor, resulting in performance degradation as compared to the pretrained policy. This phenomenon has also been reported by \citet{nair2020awac} and we provide a detailed discussion in Appendix~\ref{appendix:challenge}. In this paper, we propose ROT which addresses this issue by adaptively keeping the policy close to the behavior data during the initial phase of finetuning and reduces this dependence over time. Fig.~\ref{fig:toy}~(c) demonstrates the performance of our approach on such finetuning. It can be clearly seen that even though the BC policy is suboptimal, our proposed adaptive regularization scheme quickly improves and solves the task by driving it closer to the expert demonstration. In Fig.~\ref{fig:toy}~(d), we demonstrate that even if the agent was initialized at points outside the expert trajectory, the agent is still able to learn quickly and complete the task. This generalization to starting states would not be possible with regular BC.

\section{Regularized Optimal Transport}
\label{approach}

A fundamental challenge in imitation learning is to balance the ability to mimic demonstrated actions along with the ability to recover from states outside the distribution of demonstrated states. Behavior Cloning~(BC) specializes in mimicking demonstrated actions through supervised learning, while Inverse Reinforcement Learning~(IRL) specializes in obtaining policies that can recover from arbitrary states. Regularized Optimal Transport~(ROT) combines the best of both worlds by adaptively combining the two objectives. This is done in two phases. In the first phase, a randomly initialized policy is trained using the BC objective on expert demonstrated data. This `BC-pretrained' policy then serves as an initialization for the second phase. In the second phase, the policy is allowed access to the environment where it can train using an IRL objective. To accelerate the IRL training, the BC loss is added to the objective with an adaptive weight. Details of each component are described below, with additional algorithmic details in Appendix~\ref{appendix:alg}. 

\subsection{Phase 1: BC Pretraining} 

BC corresponds to solving the maximum likelihood problem shown in Eq.~\ref{eq:bc}. Here $\mathcal{T}^{e}$ refers to expert demonstrations. When parameterized by a normal distribution with fixed variance, the objective can be framed as a regression problem where, given inputs $s^e$, $\pi^{BC}$ needs to output $a^e$.
\begin{equation}
    \mathcal{L}^{BC} = \mathbb{E}_{(s^{e},a^{e})\sim \mathcal{T}^{e}} \|a^{e} - \pi^{BC}(s^{e})\|^{2}
    \label{eq:bc}
\end{equation}

 After training, it enables $\pi^{BC}$ to mimic the actions corresponding to the observations seen in the demonstrations. However, during rollouts in an environment, small errors in action prediction can lead to the agent visiting states not seen in the demonstrations~\cite{ross2011reduction}. This distributional mismatch often causes $\pi^{BC}$ to fail on empirical benchmarks~\cite{rajeswaran2017learning,cohen2022imitation} (see Fig.~\ref{fig:toy}~(a) in Sec.~\ref{challenge}). 

\subsection{Phase 2: Online Finetuning with IRL}
\label{subsec:finetune}

Given a pretrained $\pi^{BC}$ model, we now begin online `finetuning' of the policy $\pi^b\equiv\pi^{ROT}$ in the environment. Since we are operating without explicit task rewards, we use rewards obtained through OT-based trajectory matching, which is described in Section~\ref{background}. These OT-based rewards $r^{OT}$ enable the use of standard RL optimizers to maximize cumulative reward from $\pi^b\equiv\pi^{ROT}$. In this work we use n-step DDPG~\cite{lillicrap2015continuous}, a deterministic actor-critic based method that provides high-performance in continuous control~\cite{yarats2021mastering}. 

\paragraph{Finetuning with Regularization} $\pi^{BC}$ is susceptible to distribution shift due to accumulation of errors during online rollouts~\cite{ross2011reduction} and directly finetuning $\pi^{BC}$ also leads to subpar performance (refer to Fig.~\ref{fig:toy} in Sec.~\ref{challenge}). To address this, we build upon prior work in guided RL~\cite{rajeswaran2017learning} and offline RL~\cite{nair2020awac}, and regularize the training of $\pi^{ROT}$ by combining it with a BC loss as seen in Eq.~\ref{eq:rot}.

\begin{equation}
    \pi^{ROT} = \operatorname*{argmax}_\pi \left[(1-\lambda(\pi)))\mathbb{E}_{(s,a)\sim \mathcal{D}_{\beta}}[Q(s,a)] - \alpha \lambda(\pi)\mathbb{E}_{(s^{e},a^{e})\sim \mathcal{T}^{e}} \|a^{e} - \pi(s^{e})\|^{2} \right]
    \label{eq:rot}
\end{equation}

Here, $Q(s,a)$ represents the Q-value from the critic used in actor-critic policy optimization. $\alpha$ is a fixed weight, while $\lambda(\pi)$ is a policy-dependent adaptive weight that controls the contributions of the two loss terms. $\mathcal{D}_{\beta}$ refers to the replay buffer for online rollouts.

\paragraph{Adaptive Regularization with Soft Q-filtering} While prior work~\cite{rajeswaran2017learning, jena2020augmenting} use hand-tuned schedules for $\lambda(\pi)$, we propose a new adaptive scheme that removes the need for tuning. This is done by comparing the performance of the current policy $\pi^{ROT}$ and the pretrained policy $\pi^{BC}$ on a batch of data sampled from the replay buffer for online rollouts $\mathcal{D}_{\beta}$. More precisely, given a behavior policy $\pi^{BC}(s)$, the current policy $\pi^{ROT}(s)$, the Q-function $Q(s,a)$ and the replay buffer $\mathcal{D}_{\beta}$, we set $\lambda$ as:

\begin{equation}
    \lambda(\pi^{ROT}) = \mathbb{E}_{(s,\cdot)\sim \mathcal{D}_{\beta}}\left[\mathbbm{1}_{Q(s,\pi^{BC}(s))>Q(s,\pi^{ROT}(s))} \right]
\end{equation}
The strength of the BC regularization hence depends on the performance of the current policy with respect to the behavior policy. This filtering strategy is inspired by \citet{nair2018overcoming}, where instead of a binary hard assignment we use a soft continuous weight. Experimental comparisons with hand-tuned decay strategies are presented in Section~\ref{sec:sqf-exp}.

\paragraph{Considerations for image-based observations} Since we are interested in using ROT with high-dimensional visual observations, additional machinery is required to ensure compatibility. Following prior work in image-based RL and imitation~\cite{yarats2021mastering,cohen2022imitation}, we perform data augmentations on visual observations and then feed it into a CNN encoder. Similar to \citet{cohen2022imitation}, we use a target encoder with Polyak averaging to obtain representations for OT reward computation. This is necessary to reduce the non-stationarity caused by learning the encoder alongside the ROT imitation process.
Further implementation details and the training procedure can be found in Appendix~\ref{appendix:alg}.

\section{Experiments}
\label{experiments}
Our experiments are designed to answer the following questions: (a) How efficient is ROT for imitation learning? (b) How does ROT perform on real-world tasks? (c) How important is the choice of IRL method in ROT? (d) Does soft Q-filtering improve imitation? (e) How does ROT compare to standard reward-based RL? Additional results and analysis have been provided in Appendix~\ref{appendix:learning}.

\paragraph{Simulated tasks}
 We experiment with 10 tasks from the DeepMind Control suite~\cite{tassa2018deepmind,todorov2012mujoco}, 3 tasks from the OpenAI Robotics suite~\cite{plappert2018multi}, and 7 tasks from the Meta-world suite~\cite{yu2019meta}. For DeepMind Control tasks, we train expert policies using DrQ-v2~\cite{yarats2021mastering} and collect 10 demonstrations for each task using this policy. For OpenAI Robotics tasks, we train a state-based DrQ-v2 with hindsight experience replay~\cite{andrychowicz2017hindsight} and collect 50 demonstrations for each task. For Meta-world tasks, we use a single hard-coded expert demonstration from their open-source implementation~\cite{yu2019meta}. Full environment details can be found in Appendix~\ref{appendix:envs} and details about the variations in demonstrations and initialization conditions can be found in Appendix~\ref{appendix:demos}.

\paragraph{Robot tasks} Our real world setup for each of the 14 manipulation tasks can be seen in Fig.~\ref{fig:robot_results}. We use an Ufactory xArm 7 robot with a xArm Gripper as the robot platform for our real world experiments. However, our method is agnostic to the specific robot hardware. The observations are RGB images from a fixed camera. In this setup, we only use a single expert demonstration collected by a human operator with a joystick and limit the online training to a fixed period of 1 hour. Descriptions of each task and the evaluation procedure is in Appendix~\ref{appendix:robot_tasks}.

\begin{figure}[t!]
    \centering
    \includegraphics[width=\textwidth]{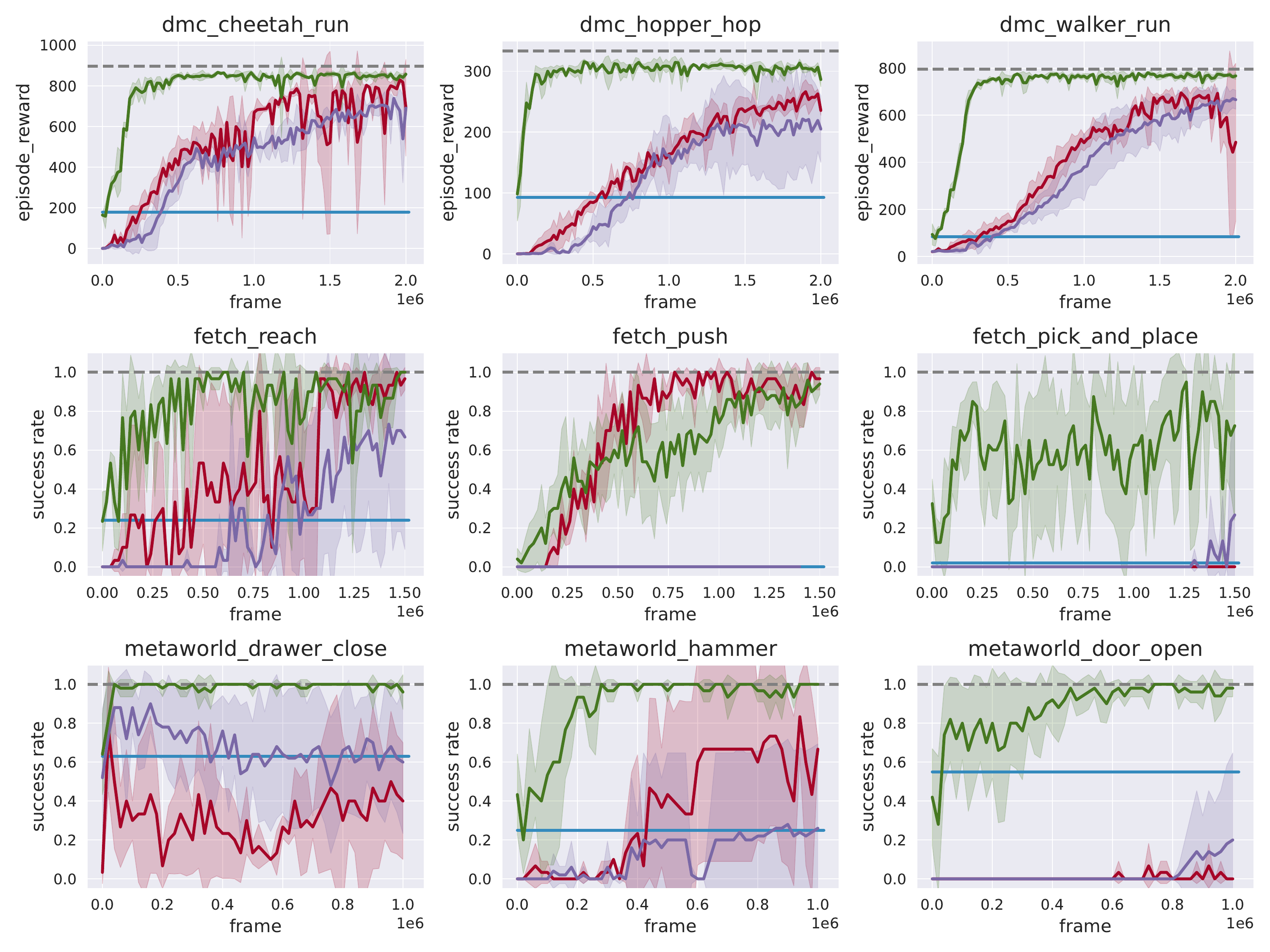}
    \cblock{128}{128}{128}\hspace{1mm}Expert\hspace{1.5mm}
    \cblock{52}{138}{189}\hspace{1mm}BC\hspace{1.5mm}
    \cblock{122}{104}{166}\hspace{1mm}OT\hspace{1.5mm}
    \cblock{166}{6}{40}\hspace{1mm}DAC\hspace{1.5mm}
    \cblock{70}{120}{33}\hspace{1mm}ROT (Ours)\hspace{1.5mm}
    \caption{Pixel-based continuous control learning on 9 selected environments. Shaded region represents $\pm1$ standard deviation across 5 seeds. We notice that ROT is significantly more sample efficient compared to prior work.}
    \label{fig:image_results}
    
\end{figure}

\paragraph{Primary baselines} We compare ROT with baselines against several prominent imitation learning methods. While a full description of our baselines are in Appendix~\ref{appendix:baselines}, a brief description of the two strongest ones are as follows:

\begin{enumerate}[leftmargin=*,align=left]
    \item \textbf{Adversarial IRL (DAC):} Discriminator Actor Critic ~\cite{kostrikov2018discriminator} is a state-of-the-art adversarial imitation learning method ~\cite{ho2016generative,torabi2018generative,kostrikov2018discriminator}. DAC outperforms prior work such as GAIL~\cite{ho2016generative} and AIRL~\cite{fu2017learning}, and thus it serves as our primary adversarial imitation baseline.
    \item \textbf{Trajectory-matching IRL (OT):} Sinkhorn Imitation Learning~\cite{papagiannis2020imitation,dadashi2020primal} is a state-of-the-art trajectory-matching imitation learning method~\cite{ghasemipour2020divergence} that approximates OT matching through the Sinkhorn Knopp algorithm~\cite{sinkhorn1967concerning,cuturi2013sinkhorn}. Since ROT is derived from similar OT-based foundations, we use SIL as our primary state-matching imitation baseline. 
\end{enumerate}

\subsection{How efficient is ROT for imitation learning?}
\label{sec:eff}
Performance of ROT for image-based imitation is depicted on select environments in Fig. \ref{fig:image_results}. On all but one task, ROT trains significantly faster than prior work. To reach $90\%$ of expert performance, ROT is on average $8.7\times$ faster on DeepMind Control tasks, $2.1\times$ faster on OpenAI Robotics tasks, and $8.9\times$ faster on Meta-world tasks. We also find that the improvements of ROT are most apparent on the harder tasks, which are in rightmost column of Fig.~\ref{fig:image_results}. 
Appendix ~\ref{appendix:eff} shows results on all 20 simulated tasks, along with experiments that exhibit similar improvements in state-based settings.

\subsection{How does ROT perform on real-world tasks?}
\begin{figure}[t!]
    \centering
    \includegraphics[width=\textwidth]{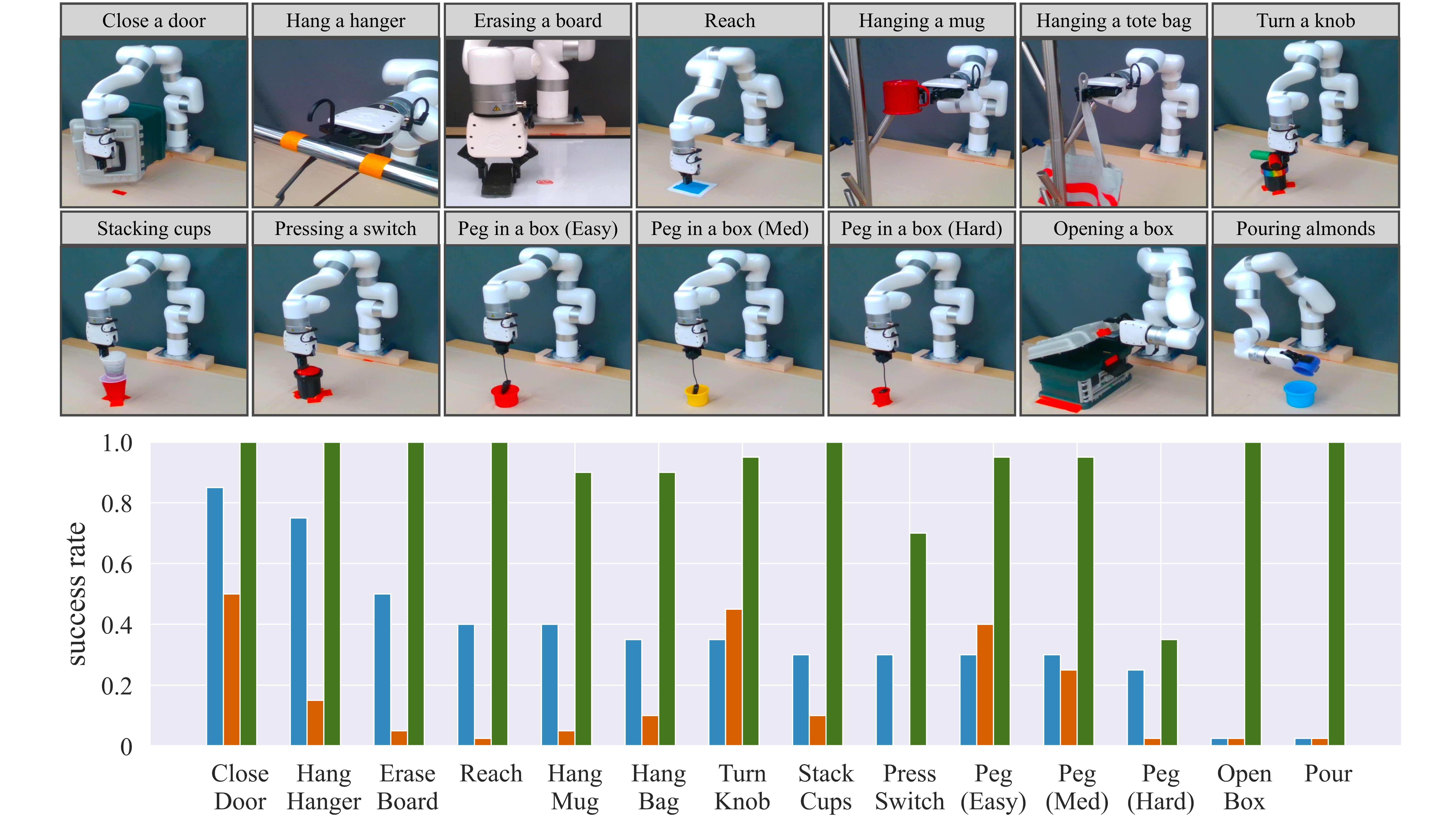}
    \cblock{52}{138}{189}\hspace{1mm}BC\hspace{1.5mm}
    \cblock{213}{94}{0}\hspace{1mm}RDAC\hspace{1.5mm}
    \cblock{70}{120}{33}\hspace{1mm}ROT (Ours)\hspace{1.5mm}
    \caption{\textbf{(Top)}~ROT is evaluated on a set of 14 robotic manipulation tasks. \textbf{(Bottom)}~Success rates for each task is computed by running 20 trajectories from varying initial conditions on the robot.}
    \label{fig:robot_results}
\end{figure}

We devise a set of 14 manipulation tasks on our xArm robot to compare the performance of ROT with BC and our strongest baseline RDAC, an adversarial IRL method that combines DAC~\cite{kostrikov2018discriminator} with our pretraining and regularization scheme. The BC policy is trained using supervised learning on a single expert demonstration collected by a human operator. ROT and RDAC finetune the pretrained BC policy through 1 hour of online training, which amounts to $\sim$ 6k environment steps. Since there is just one demonstration, our tasks are designed to have random initializations but fixed goals. Note that a single demonstration only demonstrates solving the tasks from one initial condition. Evaluation results across 20 different initial conditions can be seen in Fig.~\ref{fig:robot_results}. We observe that ROT has an average success rate of 90.1\% over 20 evaluation trajectories across all tasks as compared to 36.1\% for BC and 14.6\% for RDAC. The poor performance of BC can be attributed to distributional mismatch due to accumulation of error in online rollouts and different initial conditions. The poor performance of RDAC can be attributed to slow learning during the initial phase of training. More detailed evaluations of RDAC on simulated environments is present in Sec.~\ref{sec:choice_of_irl}.

\begin{wrapfigure}[14]{r}{0.6\textwidth}
    \vspace{-15pt}
    \centering
    \includegraphics[width=\linewidth]{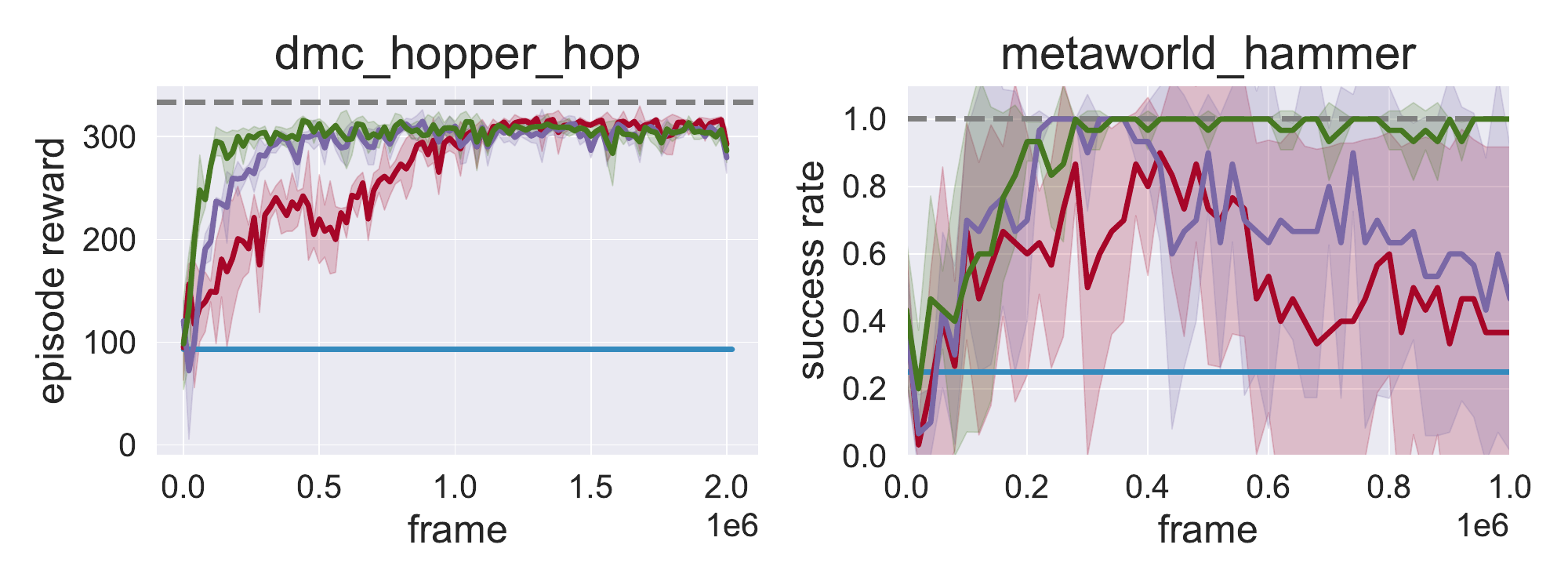}
    \hspace{6mm}\cblock{166}{6}{40}\hspace{1mm}Finetune with fixed weight\hspace{1.5mm}
    \cblock{122}{104}{166}\hspace{1mm}Finetune with fixed schedule\hspace{1.5mm}
    \cblock{70}{120}{33}\hspace{1mm}ROT (Ours)\hspace{1.5mm}
    \caption{Effect of various BC regularization schemes compared with our adaptive soft-Q filtering regularization. \vspace{20pt}}
    
    \label{fig:importance_of_sqf}
\end{wrapfigure}

\subsection{Does soft Q-filtering improve imitation?}
\label{sec:sqf-exp}
To understand the importance of soft Q-filtering, we compare ROT against two variants of our proposed regularization scheme: (a) A tuned fixed BC regularization weight (ignoring $\lambda(\pi)$ in Eq.~\ref{eq:rot}); (b) A carefully designed linear-decay schedule for $\lambda(\pi)$, where it varies from $1.0$ to $0.0$ in the first 20k environment steps~\cite{rajeswaran2017learning}. As demonstrated in Fig.~\ref{fig:importance_of_sqf} (and Appendix~\ref{sec:appendix_sqf-exp}), ROT is on par and in some cases exceeds the efficiency of a hand-tuned decay schedule, while not having to hand-tune its regularization weights. We hypothesize this improvement is primarily due to the better stability of adaptive weighing as seen in the significantly smaller standard deviation on the Meta-world tasks. 

\subsection{How important is the choice of IRL method in ROT?}
\label{sec:choice_of_irl}
\begin{figure}[t!]
    \centering
    \includegraphics[width=\textwidth]{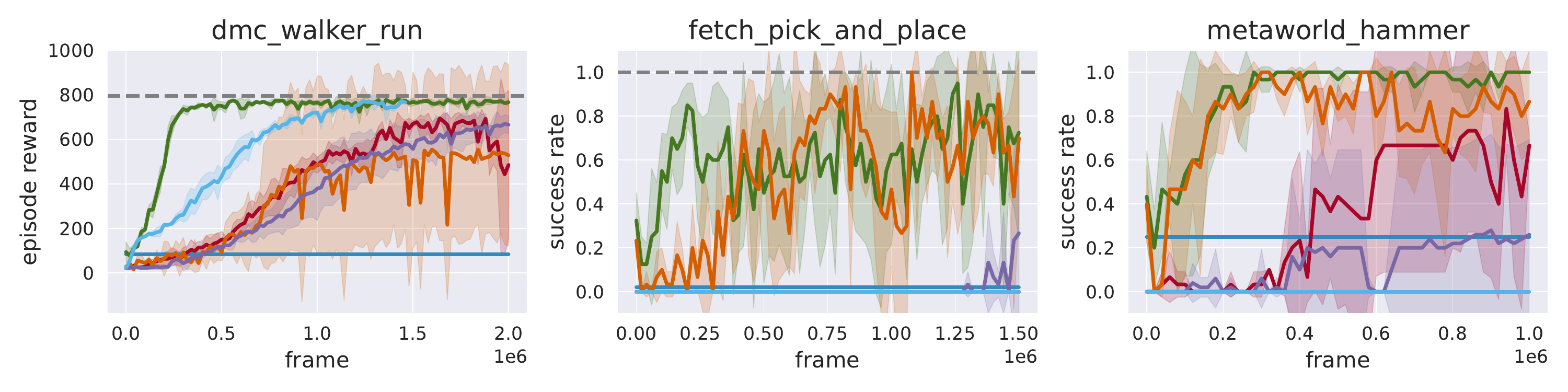}
    \cblock{128}{128}{128}\hspace{1mm}Expert\hspace{1.5mm}
    \cblock{52}{138}{189}\hspace{1mm}BC\hspace{1.5mm}
    \cblock{166}{6}{40}\hspace{1mm}DAC\hspace{1.5mm}
    \cblock{122}{104}{166}\hspace{1mm}OT\hspace{1.5mm}
    \cblock{86}{180}{233}\hspace{1mm}DrQ-v2 (RL)\hspace{1.5mm}
    \cblock{213}{94}{0}\hspace{1mm}RDAC\hspace{1.5mm}
    \cblock{70}{120}{33}\hspace{1mm}ROT (Ours)\hspace{1.5mm}
    \caption{Ablation analysis on the choice of base IRL method. We find that although adversarial methods benefit from regularized BC, the gains seen are smaller compared to ROT. Here, we also see that ROT can outperform plain RL that requires explicit task-rewards.}
    \label{fig:choice_of_irl_reward_rl}
\end{figure}

In ROT, we build on OT-based IRL instead of adversarial IRL. This is because adversarial IRL methods require iterative reward learning, which produces a highly non-stationary reward function for policy optimization. In  Fig.~\ref{fig:choice_of_irl_reward_rl}, we compare ROT with adversarial IRL methods that use our pretraining and adaptive BC regularization technique (RDAC). We find that our soft Q-filtering method does improve prior state-of-the-art adversarial IRL (RDAC vs. DAC in Fig.~\ref{fig:choice_of_irl_reward_rl}). However, our OT-based approach (ROT) is more stable and on average leads to more efficient learning.

\subsection{How does ROT compare to standard reward-based RL?}
\label{sec:reward_rl}
We compare the performance of ROT against DrQ-v2~\cite{yarats2021mastering}, a state-of-the-art algorithm for image-based RL. As opposed to the reward-free setting ROT operates in, DrQ-v2 has access to environments rewards. The results in Fig.~\ref{fig:choice_of_irl_reward_rl} show that ROT handily outperforms DrQ-v2. This clearly demonstrates the usefulness of imitation learning in domains where expert demonstrations are available over reward-based RL. We also compare against a demo-assisted variant of DrQ-v2 agent using the same pretraining and regularization scheme as ROT (refer to Appendix~\ref{appendix:reward_rl}). Interestingly, we find that our soft Q-filtering based regularization can accelerate learning of RL with task rewards, which can be seen in the high performance of the  demo-assisted variant of DrQ-v2.

\section{Related Work}
\label{related_work}

\paragraph{Imitation Learning (IL)} IL~\cite{hussein2017imitation} refers to the setting where agents learn from demonstrations without access to environment rewards. IL can be broadly categorized into Behavior Cloning (BC)~\cite{pomerleau1998autonomous,torabi2019recent,arunachalam2022dexterous,pari2021surprising} and Inverse Reinforcement Learning (IRL)~\cite{ng2000algorithms,abbeel2004apprenticeship}. BC solely learns from offline demonstrations but suffers on out-of-distributions samples~\cite{ross2011reduction} whereas IRL focuses on learning a robust reward function through online interactions but suffers from sample inefficiency~\cite{kostrikov2018discriminator}. Deep IRL methods can be further divided into two categories: $(1)$ adversarial learning~\cite{goodfellow2014generative} based methods, and $(2)$ state-matching~\cite{villani2009optimal, peyre2019computational} based methods. GAIL~\cite{ho2016generative} is an adversarial learning based formulation inspired by maximum entropy IRL~\cite{ziebart2008maximum} and GANs~\cite{goodfellow2014generative}. There has been a significant body of work built up on GAIL proposing alternative losses~\cite{fu2017learning,xiao2019wasserstein,torabi2018generative}, and enhancing its sample efficiency by porting it to an off-policy setting~\cite{kostrikov2018discriminator}. There have also been visual extensions of these adversarial learning approaches~\cite{cetin2021domain,toyer2020magical,rafailov2021visual,cohen2022imitation}. However, although adversarial methods produce competent policies, they are inefficient due to the non-stationarity associated with iterative reward inference~\cite{cohen2022imitation}.

\paragraph{Optimal Transport (OT)} OT~\cite{villani2009optimal,peyre2019computational} is a tool for comparing probability measures while including the geometry of the space. In the context of IL, OT computes an alignment between a set of agent and expert observations using distance metrics such as Sinkhorn~\cite{cuturi2013sinkhorn}, Gromov-Wasserstein~\cite{peyre2016gromov}, GDTW~\cite{cohen2021aligning}, CO-OT~\cite{redko2020co} and Soft-DTW~\cite{cuturi2017soft}. For many of these distance measures, there is an associated IL algorithm, with SIL~\cite{papagiannis2020imitation} using Sinkhorn, PWIL~\cite{dadashi2020primal} using greedy Wasserstein, GDTW-IL~\cite{cohen2021aligning} using GDTW, and GWIL~\cite{fickinger2021cross} using Gromov-Wasserstein. Recent work from \citet{cohen2022imitation} demonstrates that the Sinkhorn distance~\cite{papagiannis2020imitation} produces the most efficient learning among the discussed metrics. They further show that SIL is compatible with high-dimensional visual observations and encoded representations. Inspired by this, ROT adopts the Sinkhorn metric for its OT reward computation, and improves upon SIL through adaptive behavior regularization.

\paragraph{Behavior Regularized Control} Behavior regularization is a widely used technique in offline RL~\cite{levine2020offline} where explicit constraints are added to the policy improvement update to avoid bootstrapping on out-of-distribution actions~\cite{fujimoto2021minimalist,wu2019behavior,ajay2021opal, kumar2019stabilizing,siegel2020keep,fujimoto2019off}. In an online setting with access to environment rewards, prior work~\cite{rajeswaran2017learning,uchendu2022jump} has shown that behavior regularization can be used to boost sample efficiency by finetuning a pretrained policy via online interactions. For instance, \citet{jena2020augmenting} demonstrates the effectiveness of behavior regularization to enhance sample efficiency in the context of adversarial IL. ROT builds upon this idea by extending to visual observations, OT-based IL, and adaptive regularization, which leads to improved performance (see Appendix~\ref{appendix:ablations}). We also note that the idea of using adaptive regularization has been previously explored in RL~\cite{nair2018overcoming}. However, ROT uses a soft, continuous adaptive scheme, which on initial experiments provided significantly faster learning compared to hard assignments.

\section{Conclusion and Limitations}
\label{limitations}

In this work, we have proposed a new imitation learning algorithm, ROT, that demonstrates improved performance compared to prior state-of-the-art work on a variety of simulated and robotic domains.
However, we recognize a few limitations in this work: (a) Since our OT-based approach aligns agents with demonstrations without task-specific rewards, it relies on the demonstrator being an `expert'. Extending ROT to suboptimal, noisy and multimodal demonstrations would be an exciting problem to tackle. (b) Performing BC pretraining and BC-based regularization requires access to expert actions, which may not be present in some real-world scenarios particularly when learning from humans. Recent work on using inverse models to infer actions given observational data could alleviate this challenge~\cite{radosavovic2020state}. (c) On robotic tasks such as \textit{Peg in box (hard)} and \textit{Pressing a switch} from Fig.~\ref{fig:robot_results}, we find that ROT's performance drops substantially compared to other tasks. This might be due to the lack of visual features corresponding to the task success. For example, in the `Peg' task, it is visually difficult to discriminate if the peg is in the box or behind the box. Similarly for the `Switch' task, it is difficult to discern if the button was pressed or not. This limitation can be addressed by integrating more sensory modalities such as additional cameras, and tactile sensors in the observation space.


\acknowledgments{We thank Ben Evans, Anthony Chen, Ulyana Piterbarg and Abitha Thankaraj for valuable feedback and discussions. This work was supported by grants from Honda, Amazon, and ONR awards N00014-21-1-2404 and N00014-21-1-2758.}


\bibliography{references}  


\newpage
\appendix

\label{sec:appendix}

\section{Background}
\label{appendix:background}

\paragraph{Reinforcement Learning (RL)}We study RL as a discounted infinite-horizon Markov Decision Process (MDP)~\cite{bellman1957markovian, sutton2018reinforcement}. For pixel observations, the agent's state is approximated as a stack of consecutive RGB frames~\cite{mnih2015human}. The MDP is of the form $(\mathcal{O}, \mathcal{A}, P, R, \gamma, d_{0})$ where $\mathcal{O}$ is the observation space, $\mathcal{A}$ is the action space, $P: \mathcal{O}\times\mathcal{A}\rightarrow\Delta(\mathcal{O})$ is the transition function that defines the probability distribution over the next state given the current state and action, $R:\mathcal{O}\times\mathcal{A}\rightarrow\mathbb{R}$ is the reward function, $\gamma$ is the discount factor and $d_{0}$ is the initial state distribution. The goal is to find a policy $\pi:\mathcal{O}\rightarrow\Delta(\mathcal{A})$ that maximizes the expected discount sum of rewards $\mathbb{E}_{\pi}[\Sigma_{t=0}^{\infty}\gamma^{t}R(\boldsymbol{o}_{t},\boldsymbol{a}_{t})]$, where $\boldsymbol{o}_{0}\sim d_{0}$, $a_{t}\sim \pi(\boldsymbol{o}_{t})$ and $\boldsymbol{o}_{t+1}\sim P(.|\boldsymbol{o}_{t},\boldsymbol{a}_{t})$.

\paragraph{Imitation Learning (IL)} The goal of imitation learning is to learn a behavior policy $\pi^b$ given access to either the expert policy $\pi^e$ or trajectories derived from the expert policy $\mathcal{T}^e$. While there are a multitude of settings with differing levels of access to the expert~\cite{torabi2019recent}, this work operates in the setting where the agent only has access to observation-based trajectories, i.e. $\mathcal{T}^e \equiv \{(o_t, a_t)_{t=0}^{T}\}_{n=0}^N$. Here $N$ and $T$ denotes the number of trajectory rollouts and episode timesteps respectively. We choose this specific setting since obtaining observations and actions from expert or near-expert demonstrators is feasible in real-world settings~\cite{zhan2020framework,young2020visual} and falls in line with recent work in this area~\cite{dadashi2020primal,ho2016generative,kostrikov2018discriminator}.

\paragraph{Inverse Reinforcement Learning~(IRL)} IRL~\cite{ng2000algorithms,abbeel2004apprenticeship} tackles the IL problem by inferring the reward function $r^e$ based on expert trajectories $\mathcal{T}^e$. Then given the inferred reward $r^e$, policy optimization is used to derive the behavior policy $\pi^b$. Prominent algorithms in IRL~\cite{kostrikov2018discriminator,ho2016generative} requires alternating the inference of reward and optimization of policy in an iterative manner, which is practical for restricted model classes~\cite{abbeel2004apprenticeship}. For compatibility with more expressive deep networks, techniques such as adversarial learning~\cite{ho2016generative,kostrikov2018discriminator} or optimal-transport~\cite{papagiannis2020imitation,dadashi2020primal,cohen2022imitation} are needed. Adversarial learning based approaches tackle this problem by learning a discriminator that models the gap between the expert trajectories $\mathcal{T}^e$ and behavior trajectories $\mathcal{T}^b$. The behavior policy $\pi^b$ is then optimized to minimize this gap through gap-minimizing rewards $r^e$. Such a training procedure is prone to instabilities since $r^e$ is updated at every iteration and is hence non-stationary for the optimization of $\pi^b$.

\paragraph{Optimal Transport for Imitation Learning~(OT)} To alleviate the non-stationary reward problem with adversarial IRL frameworks, a new line of OT-based approaches have been recently proposed~\cite{papagiannis2020imitation,dadashi2020primal,cohen2022imitation}. Intuitively, the closeness between expert trajectories $\mathcal{T}^e$ and behavior trajectories $\mathcal{T}^b$ can be computed by measuring the optimal transport of probability mass from $\mathcal{T}^b \rightarrow \mathcal{T}^e$. During policy learning, the policy $\pi_{\phi}$ encompasses a feature preprocessor $f_{\phi}$ which transforms observations into informative state representations. Some examples of a preprocessor function $f_{\phi}$ are an identity function, a mean-variance scaling function and a parametric neural network. In this work, we use a parametric neural network as $f_{\phi}$. Given a cost function $c:\mathcal{O}\times\mathcal{O}\rightarrow\mathbb{R}$ defined in the preprocessor's output space and an OT objective $g$, the optimal alignment between an expert trajectory $\textbf{o}^{e}$ and a behavior trajectory $\textbf{o}^{b}$  can be computed as 

\begin{equation}
    \mu^{*} \in \underset{\mu\in\mathcal{M}}{\text{arg~min}}~g(\mu, f_{\phi}(\textbf{o}^{b}), f_{\phi}(\textbf{o}^{e}), c)
    \label{appendix:eq:alignment}
\end{equation}

where $\mathcal{M}=\{\mu\in\mathbb{R}^{T\times T}:\mu\boldsymbol{1}=\mu^{T}\boldsymbol{1}=\frac{1}{T}\boldsymbol{1}\}$ is the set of coupling matrices and the cost $c$ can be the Euclidean or Cosine distance. In this work, inspired by \cite{cohen2022imitation}, we use the entropic Wasserstein distance with cosine cost as our OT metric, which is given by the equation

\begin{equation}
\begin{aligned}
    g(\mu, f_{\phi}(\textbf{o}^{b}), f_{\phi}(\textbf{o}^{e}), c) &=  \mathcal{W}^{2}(f_{\phi}(\textbf{o}^{b}), f_{\phi}(\textbf{o}^{e}))\\
    &= \sum_{t,t'=1}^{T}C_{t,t^{'}} \mu_{t,t'}
\end{aligned}
\label{appendix:eq:wasserstein}
\end{equation}

where the cost matrix $C_{t,t^{'}} = c(f_{\phi}(\textbf{o}^{b}), f_{\phi}(\textbf{o}^{e}))$. Using Eq.~\ref{appendix:eq:wasserstein} and the optimal alignment $\mu^{*}$ obtained by optimizing Eq.~\ref{appendix:eq:alignment}, a reward signal can be computed for each observation using the equation
\begin{equation}
    \label{appendix:eq:ot_reward}
    r^{OT}(o^{\boldsymbol{b}}_{t}) = - \sum_{t'=1}^{T} C_{t,t^{'}} \mu^{*}_{t,t^{'}}
\end{equation}

Intuitively, maximizing this reward encourages the imitating agent to produce trajectories that closely match demonstrated trajectories. Since solving Eq.~\ref{appendix:eq:alignment} is computationally expensive, approximate solutions such as the Sinkhorn algorithm~\cite{knight2008sinkhorn,papagiannis2020imitation} are used instead.

\section{Issue with Fine-tuning Actor-Critic Frameworks}
\label{appendix:challenge}
In this paper, we use $n$-step DDPG proposed by \citet{yarats2021mastering} as our RL optimizer for actor-critic based reward maximization. DDPG~\cite{lillicrap2015continuous} concurrently learns a deterministic policy $\pi_\phi$ using deterministic policy gradients (DPG)~\cite{silver2014deterministic} and a Q-function $Q_\theta$ by minimizing a n-step Bellman residual (for n-step DDPG). For a parameterized actor network $\pi_{\phi}(s)$ and a critic function $Q_{\theta}(s,a)$, the deterministic policy gradients (DPG) for updating the actor weights is given by

\begin{equation}
    \label{eq:issue_ft}
    \begin{aligned}
        \nabla_{\phi} J & \approx \mathbb{E}_{s_{t}\sim\rho_{\beta}} \left[\nabla_{\phi} \left. Q_{\theta}(s,a)\right|_{s=s_{t}, a=\pi_{\phi}(s_{t})} \right]\\
        & = \mathbb{E}_{s_{t}\sim\rho_{\beta}} \left[\nabla_{a} \left. Q_{\theta}(s,a)\right|_{s=s_{t}, a=\pi_{\phi}(s_{t})} \nabla_{\phi} \left. \pi_{\phi}(s)\right|_{s=s_{t}} \right]
    \end{aligned}
\end{equation}

Here, $\rho_{\beta}$ refers to the state visitation distribution of the data present in the replay buffer at time $t$. From Eq.~\ref{eq:issue_ft}, it is clear that the policy gradients in this framework depend on the gradients with respect to the critic value. Hence, as mentioned in \cite{nair2020awac, uchendu2022jump}, naively initializing the actor with a pretrained policy while using a randomly initialized critic results in the untrained critic providing an exceedingly poor signal to the actor network during training. As a result, the actor performance drops immediately and the good behavior of the informed initialization of the policy gets forgotten. In this paper, we propose an adaptive regularization scheme that permits finetuning a pretrained actor policy in an actor-critic framework. As opposed to ~\citet{rajeswaran2017learning, jena2020augmenting} which employ on-policy learning, our method is off-policy and aims to leverage the sample efficient characteristic of off-policy learning as compared to on-policy learning~\cite{kostrikov2018discriminator}.

\section{Algorithmic Details}
\label{appendix:alg}

\begin{algorithm}[h!]
\caption{ROT: Regularized Optimal Transport}\label{alg:rot}
\begin{algorithmic}
\State $\textbf{Require:}$\\
Expert Demonstrations $\mathcal{T}^e \equiv \{(o_t, a_t)_{t=0}^{T}\}_{n=0}^N$\\
Pretrained policy $\pi^{BC}$\\
Replay buffer $\mathcal{D}$, Training steps $T$, Episode Length $L$\\
Task environment $env$\\
Parametric networks for RL backbone (e.g., the encoder, policy and critic function for DrQ-v2)\\
A discriminator $D$ for adversarial baselines
\\
\State $\textbf{Algorithm:}$
\State $\pi^{ROT} \gets \pi^{BC}$ \Comment{Initialize with pretrained policy}
\For{each timestep $t$ = $1...T$}
    \If{done}
        \State $r_{1:L} = \text{rewarder}_{OT}\text{(episode)}$ \Comment{OT-based reward computation}
        \State $\text{Update episode with}~r_{1:L}~\text{and add}~(\textbf{o}_{t}, \textbf{a}_{t}, \textbf{o}_{t+1}, r_{t})~\text{to}~\mathcal{D}$
        \State $\textbf{o}_{t} = env.reset(),~\text{done} = \text{False},~\text{episode} = [~]$ 
    \EndIf
    \State $\textbf{a}_{t} = \pi^{ROT}(\textbf{o}_{t})$
    \State $\textbf{o}_{t+1},~\text{done} = env.step(\textbf{a}_{t})$
    \State $\text{episode.append}([\textbf{o}_{t}, \textbf{a}_{t}, \textbf{o}_{t+1}])$
    \State $\text{Update backbone-specific networks and reward-specific networks using}~\mathcal{D}$
\EndFor
\end{algorithmic}
\end{algorithm}

\subsection{Implementation}
Algorithm~\ref{alg:rot} describes our proposed algorithm, Regularized Optimal Transport (ROT), for sample efficient imitation learning for continuous control tasks. Further implementation details are as follows:

\paragraph{Algorithm and training procedure} Our model consists of 3 primary neural networks - the encoder, the actor and the critic. During the BC pretraining phase, the encoder and the actor are trained using a mean squared error (MSE) on the expert demonstrations. Next, for finetuning, weights of the pretrained encoder and actor are loaded from memory and the critic is initialized randomly. We observed that the performance of the algorithm is not very sensitive to the value of $\alpha$ and we set it to $0.03$ for all experiments in this paper. A copy of the pretrained encoder and actor are stored with fixed weights to be used for computing $\lambda(\pi)$ for soft Q-filtering.

\paragraph{Actor-critic based reward maximization} We use a recent n-step DDPG proposed by \citet{yarats2021mastering} as our RL backbone. The deterministic actor is trained using deterministic policy gradients (DPG)~\cite{silver2014deterministic} given by Eq.~\ref{eq:issue_ft}. The critic is trained using clipped double Q-learning similar to ~\citet{yarats2021mastering} in order to reduce the overestimation bias in the target value. This is done using two Q-functions, $Q_{\theta1}$ and $Q_{\theta2}$. The critic loss for each critic is given by the equation

\begin{equation}
    \mathcal{L}_{\theta_{k}} = \mathbbm{E}_{(s,a)\sim D_{\beta}} \left[(Q_{\theta_{k}}(s,a) - y)^{2} \right] \forall ~k \in \{1,2\}
\end{equation}

where $\mathcal{D}_{\beta}$ is the replay buffer for online rollouts and $y$ is the target value for n-step DDPG given by

\begin{equation}
    y = \sum_{i=0}^{n-1} \gamma^{i}r_{t+i} + \gamma^{n} \underset{k=1,2}{min}Q_{\bar \theta_{k}}(s_{t+n}, a_{t+n})
\end{equation}

Here, $\gamma$ is the discount factor, $r$ is the reward obtained using OT-based reward computation and $\bar \theta_{1}$, $\bar \theta_{2}$ are the slow moving weights of target Q-networks. 

\paragraph{Target feature processor to stabilize OT rewards} The OT rewards are computed on the output of the feature processor $f_{\phi}$ which is initialized with a parametric neural network. Hence, as the weights of $f_{\phi}$ change during training, the rewards become non-stationary resulting in unstable training. In order to increase the stability of training, the OT rewards are computed using a target feature processor $f_{\phi^{'}}$~\cite{cohen2022imitation} which is updated with the weights of $f_{\phi}$ every $T_{update}$ environment steps. For state-based observations, $f_{\phi}$ corresponds to a 'trunk' network which is a single layer neural network. For pixel-based observations, $f_{\phi}$ includes DrQ-v2's encoder followed by the 'trunk' network.

\subsection{Hyperparameters}
The complete list of hyperparameters is provided in Table~\ref{tab:hyperparams}. Similar to \citet{yarats2021mastering}, there is a slight deviation from the given setting for the Walker Stand/Walk/Run task from the DeepMind Control suite where we use a mini-batch size of 512 and a $n$-step return of 1.\\

\begin{table}[h!]
    \begin{center}
    \setlength{\tabcolsep}{18pt}
    \renewcommand{\arraystretch}{1.5}
    \begin{tabular}{ c c c } 
        \hline
        Method & Parameter & Value \\
        \hline
        Common & Replay buffer size & 150000 \\
               & Learning rate      & $1e^{-4}$\\
               & Discount $\gamma$   & 0.99\\
               & $n$-step returns   & 3\\
               & Action repeat      & 2\\
               & Seed frames        & 12000\\
               & Mini-batch size    & 256\\
               & Agent update frequency & 2\\
               & Critic soft-update rate & 0.01\\
               & Feature dim        & 50\\
               & Hidden dim         & 1024\\
               & Optimizer          & Adam\\
        \hline
        ROT    & Exploration steps   & 0\\
               & DDPG exploration schedule & 0.1\\
               & Target feature processor update frequency(steps) & 20000\\
               & Reward scale factor & 10\\
               & Fixed weight $\alpha$       & 0.03\\
               & Linear decay schedule for $\lambda(\pi)$ & linear(1,0.1,20000)\\
        \hline
        OT     & Exploration steps   & 2000\\
               & DDPG exploration schedule & linear(1,0.1,500000)\\
               & Target feature processor update frequency(steps) & 20000\\
               & Reward scale factor & 10\\
        \hline
        DAC    & Exploration steps   & 2000\\
               & DDPG exploration schedule & linear(1,0.1,500000)\\
               & Gradient penalty coefficient & 10\\
        \hline
    \end{tabular}
    \end{center}
    \caption{List of hyperparameters.}
    \label{tab:hyperparams}
\end{table}

\section{Environments}
\label{appendix:envs}
Table~\ref{tab:tasks} lists the different tasks that we experiment with from the DeepMind Control suite~\cite{tassa2018deepmind,todorov2012mujoco}, OpenAI Robotics suite~\cite{plappert2018multi} and the Meta-world suite~\cite{yu2019meta} along with the number of training steps and the number of demonstrations used. For the tasks in the OpenAI Robotics suite, we fix the goal while keeping the initial state randomized. No modifications are made in case of the DeepMind Control suite and the Meta-world suite. The episode length for all tasks in DeepMind Control is 1000 steps, for OpenAI Robotics is 50 steps and Meta-world is 125 steps (except bin picking which runs for 175 steps).

\begin{table}[h!]
    \begin{center}
    \setlength{\tabcolsep}{5pt}
    \renewcommand{\arraystretch}{1.5}
    \begin{tabular}{ c c c c } 
     \hline
     Suite & Tasks & Allowed Steps & \# Demonstrations \\
     \hline
     DeepMind Control &  Acrobot Swingup & $2\times10^{6}$ & 10 \\ 
                            & Cartpole Swingup &                 &\\
                            & Cheetah Run      &                 &\\
                            & Finger Spin      &                 &\\
                            & Hopper Stand     &                 &\\
                            & Hopper Hop       &                 &\\
                            & Quadruped Run    &                 &\\
                            & Walker Stand     &                 &\\
                            & Walker Walk      &                 &\\
                            & Walker Run       &                 &\\
     \hline
     OpenAI Robotics & Fetch Reach          & $1.5\times10^{6}$ & 50\\ 
                           & Fetch Push           &                   & \\
                           & Fetch Pick and Place &                   & \\
     \hline
     Meta-World           & Hammer                 & $1\times10^{6}$ & 1\\ 
                          & Drawer Close           &                 &\\
                          & Door Open              &                 &\\
                          & Bin Picking            &                 &\\
                          & Button Press Topdown   &                 &\\
                          & Door Unlock.           &                 &\\
     \hline
     xArm Robot           & Close Door             & $6\times10^{3}$ & 1\\ 
                          & Hang Hanger            &                 &\\
                          & Erase Board            &                 &\\
                          & Reach                  &                 &\\
                          & Hang Mug               &                 &\\
                          & Hang Bag               &                 &\\
                          & Turn Knob              &                 &\\
                          & Stack Cups             &                 &\\
                          & Press Switch           &                 &\\
                          & Peg (Easy)             &                 &\\
                          & Peg (Medium)           &                 &\\
                          & Peg (Hard)             &                 &\\
                          & Open Box               &                 &\\
                          & Pour                   &                 &\\
     \hline
    \end{tabular}
    \end{center}
    \caption{List of tasks used for evaluation.}
    \label{tab:tasks}
\end{table}

\section{Demonstrations}
\label{appendix:demos}
For DeepMind Control tasks, we train expert policies using pixel-based DrQ-v2~\cite{yarats2021mastering} and collect 10 demonstrations for each task using this expert policy. The expert policy is trained using a stack of $3$ consecutive RGB frames of size $84\times84$ with random crop augmentation. Each action in the environment is repeated 2 times. For OpenAI Robotics tasks, we train a state-based DrQ-v2 with hindsight experience replay~\cite{andrychowicz2017hindsight} and collect 50 demonstrations for each task. The state representation comprises the observation from the environment appended with the desired goal location. For this, we did not do frame stacking and action repeat was set to 2. For Meta-World tasks, we use a single expert demonstration obtained using the task-specific hard-coded policies provided in their open-source implementation~\cite{yu2019meta}.

\begin{figure}[t!]
    \centering
    \includegraphics[width=\textwidth]{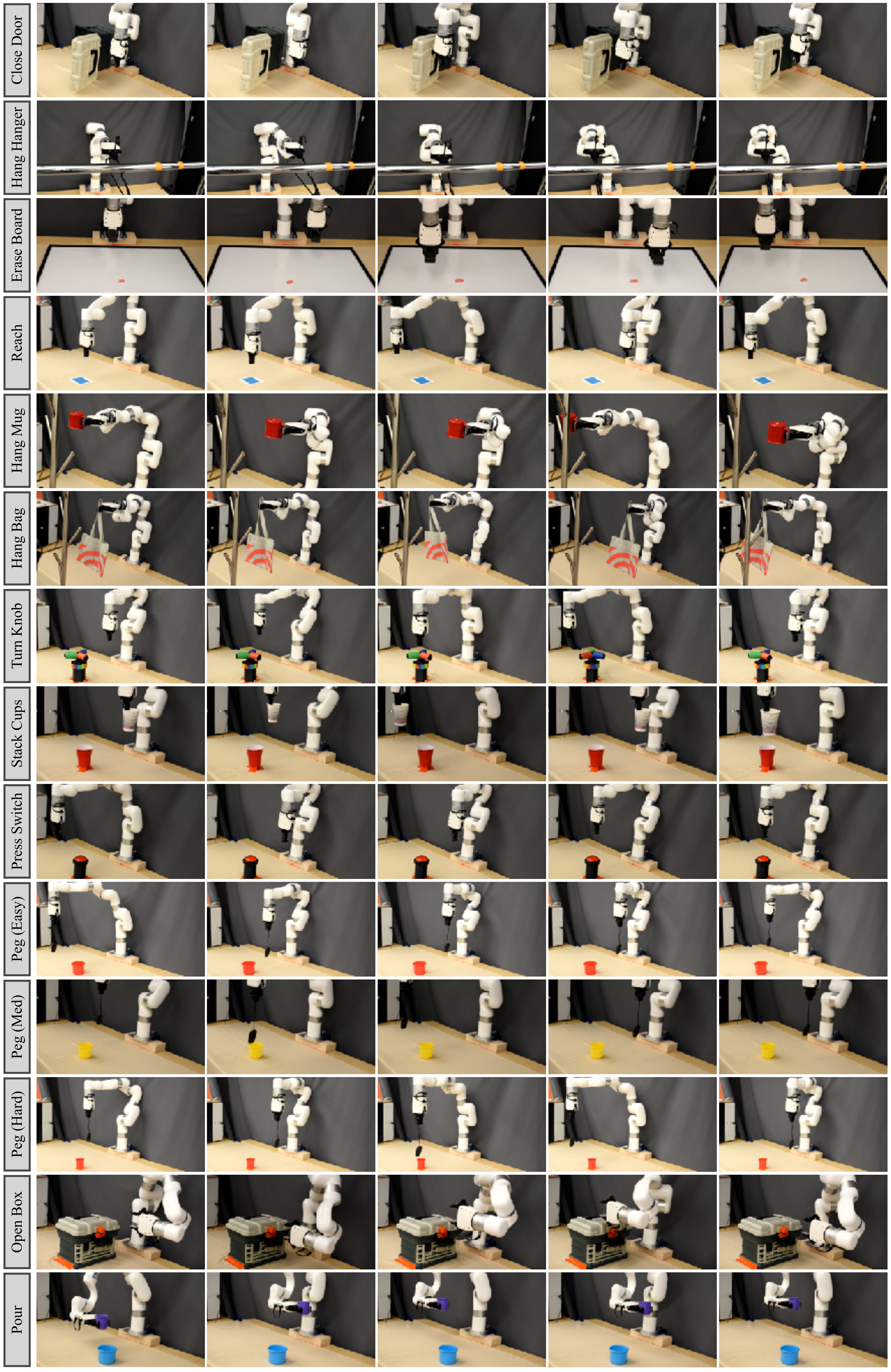}
    \caption{Examples of different initializations for the real robot tasks.}
    \label{fig:appendix_diff_init_robot}
\end{figure}

\section{Robot Tasks}
\label{appendix:robot_tasks}
In this section, we describe the suite of manipulation experiments carried out on a real robot in this paper.

\begin{enumerate}[label=(\alph*),leftmargin=*]
    \item \textbf{Door Close:} Here, the robot arm is supposed to close an open door by pushing it to the target.
    \item \textbf{Hang Hanger:} While holding a hanger between the grippers, the robot arm is initialized at random position and is tasked with putting the hanger at a goal region on a closet rod. 
    \item \textbf{Erase Board:} While holding a board duster between the grippers, the robot arm is tasks with erasing marking drawn on the board while getting initialized from random positions.
    \item \textbf{Reach:} The robot arm is required to reach a specific goal after being initialized at a random position.
    \item \textbf{Hang Mug:} While holding a mug between the grippers, the robot arm is initialized at random position and is tasked with hanging the mug on a specific hook.
    \item \textbf{Hang Bag:} While holding a tote between the grippers, the robot arm is initialized at random position and is tasked with hanging the tote bag on a specific hook.
    \item \textbf{Turn Knob:} The robot arm is tasked with rotating a knob placed on the table by a certain angle after being initialized at a random position. We consider a 90 degree rotation as success.
    \item \textbf{Cup Stack:} While holding a cup between the gripper, the robot arm is required to stack it into another cup placed on the table.
    \item \textbf{Press Switch:} With the gripper kept closed, the robot arm is required to press a switch (with an LED light) placed on the table.
    \item \textbf{Peg (Easy, Medium, Hard):} The robot arm is supposed to insert a peg, hanging by a string, into a bucket placed on the table. This task has 3 variants - Easy, Medium, Hard - with the size of the bucket decreasing from Easy to Hard.
    \item \textbf{Box Open:} In this task, the robot arm is supposed to open the lid of a box placed on the table by lifting a handle provided in the front of the box. 
    \item \textbf{Pour:} Given a cup with some item place inside (in our case, almonds), the robot arm is supposed to move towards a cup place on the table and pour the item into the cup.
\end{enumerate}

\paragraph{Evaluation procedure} For each task, we obtained a set of 20 random initializations and evaluate all of the methods (BC, RDAC and ROT) over 20 trajectories from the same set of initializations. These initializations are different for each task based on the limits of the observation space for the task.

\begin{figure}[t!]
    \centering
    \includegraphics[width=\textwidth]{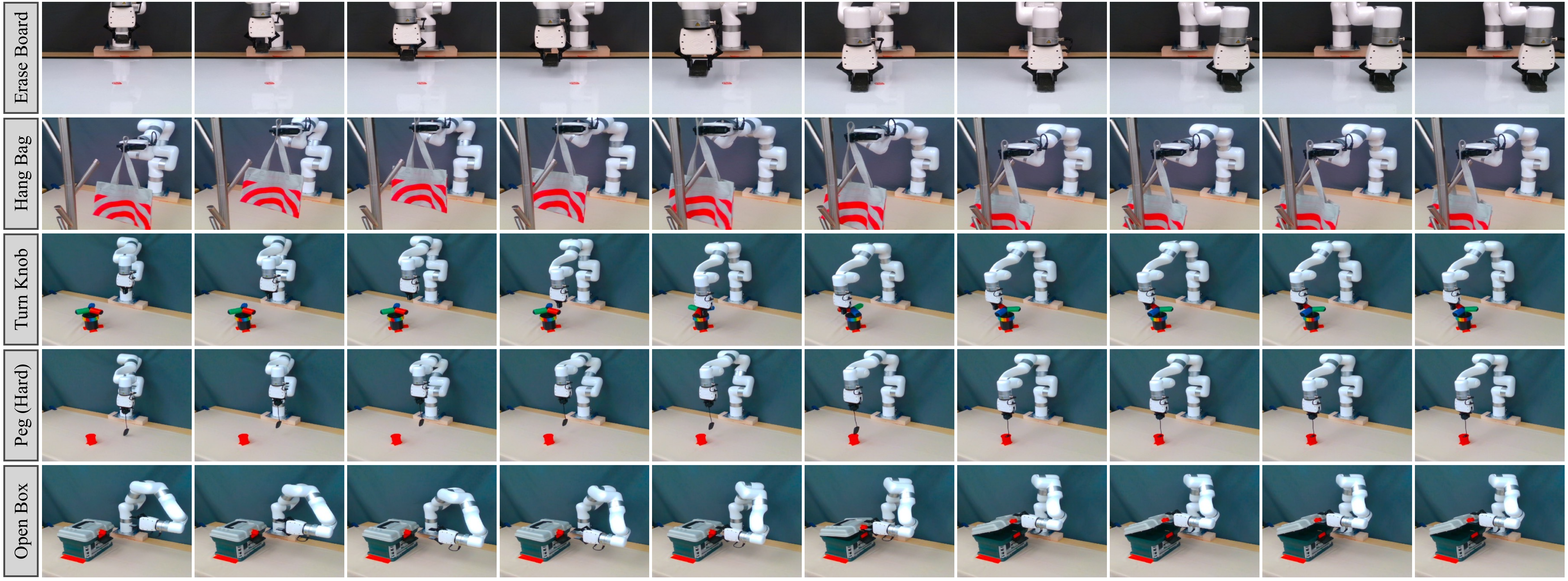}
    \caption{Example trajectories for selected real robot tasks.}
    \label{fig:appendix_trajectory_example}
\end{figure}

\section{Baselines}
\label{appendix:baselines}
Throughout the paper, we compare ROT with several prominent imitation learning and reinforcement learning methods. Here, we give a brief description of each of the baseline models that have been used.

\begin{enumerate}[label=(\alph*),leftmargin=*]
    \item \textbf{Expert:} For each task, the expert refers to the expert policy used to generate the demonstrations for the task (described in Appendix~\ref{appendix:demos}). 
    \item \textbf{Behavior Cloning (BC):} This refers to the behavior cloned policy trained on expert demonstrations.
    \item \textbf{Adversarial IRL (DAC):} Discriminator Actor Critic ~\cite{kostrikov2018discriminator} is a state-of-the-art adversarial imitation learning method ~\cite{ho2016generative,torabi2018generative,kostrikov2018discriminator}. Since DAC outperforms prior work such as GAIL\cite{ho2016generative} and AIRL\cite{fu2017learning}, it serves as our primary adversarial imitation baseline.
    \item \textbf{State-matching IRL (OT):} Sinkhorn Imitation Learning~\cite{papagiannis2020imitation,dadashi2020primal} is a state-of-the-art state-matching imitation learning method~\cite{ghasemipour2020divergence} that approximates OT matching through the Sinkhorn Knopp algorithm. Since ROT is derived from similar OT-based foundations, we use SIL as our primary state-matching imitation baseline. 
    \item \textbf{RDAC:} This is the same as ROT, but instead of using state-matching IRL (OT), adversarial IRL (DAC) is used.
    \item \textbf{Finetune with fixed weight:} This is similar to ROT where instead of using a time-varying adaptive weight $\lambda(i)$, only the fixed weight $\lambda_{0}$ is used. $\lambda_{0}$ is set to a fixed value of $0.03$.
    \item \textbf{Finetune with fixed schedule:} This is similar to ROT  that uses both the fixed weight $\lambda_{0}$ and the time-varying adaptive weight $\lambda_{1}(i)$. However, instead of using Soft Q-filtering to compute $\lambda_{1}(i)$, a hand-coded linear decay schedule is used.
    \item \textbf{DrQ-v2 (RL):} DrQ-v2~\cite{yarats2021mastering} is a state-of-the-art algorithm for pixel-based RL. DrQ-v2 is assumed to have access to environment rewards as opposed to ROT which computes the reward using OT-based techniques.
    \item \textbf{Demo-DrQ-v2:} This refers to DrQ-v2 but with access to both environment rewards and expert demonstrations. The model is initialized with a pretrained BC policy followed by RL finetuning with an adaptive regularization scheme like ROT. During RL finetuning, this baseline has access to environment rewards.
    \item \textbf{BC+OT:} This is the same as the OT baseline but the policy is initialized with a pretrained BC policy. No adaptive regularization scheme is used while finetuning the pretrained policy. 
    \item \textbf{OT+BC Reg.:} This is the same as the OT baseline with randomly initialized networks but during training, the adaptive regularization scheme is added to the objective function.
\end{enumerate}

\section{Additional Experimental Results}
\label{appendix:learning}

\subsection{How efficient is ROT for imitation learning?}
\label{appendix:eff}
\begin{figure}[t!]
    \centering
    \includegraphics[width=\textwidth]{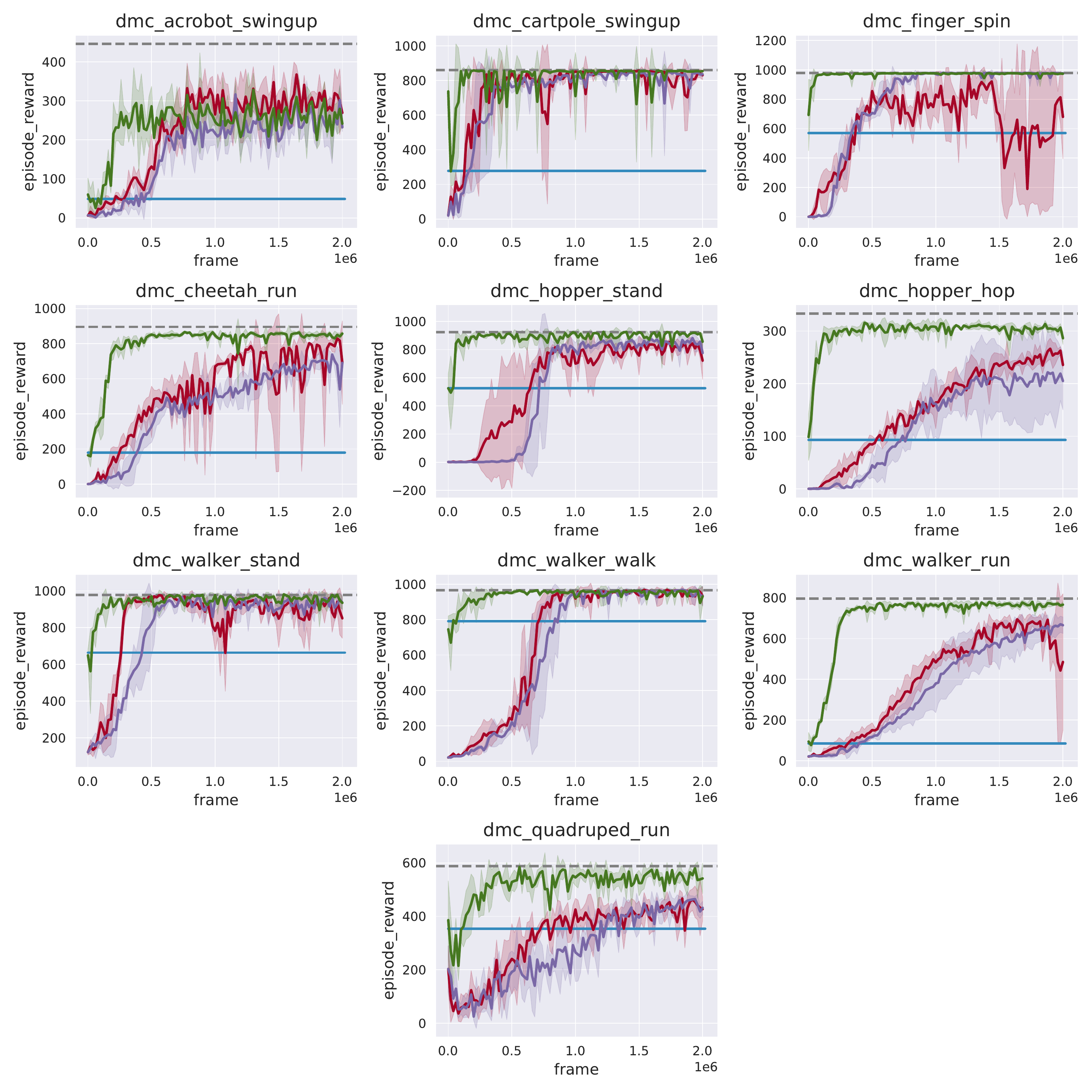}
    \cblock{128}{128}{128}\hspace{1mm}Expert\hspace{1.5mm}
    \cblock{52}{138}{189}\hspace{1mm}BC\hspace{1.5mm}
    \cblock{122}{104}{166}\hspace{1mm}OT\hspace{1.5mm}
    \cblock{166}{6}{40}\hspace{1mm}DAC\hspace{1.5mm}
    \cblock{70}{120}{33}\hspace{1mm}ROT (Ours)\hspace{1.5mm}
    \caption{Pixel-based continuous control learning on 10 DMC environments. Shaded region represents $\pm1$ standard deviation across 5 seeds. We notice that ROT is significantly more sample efficient compared to prior work.}
    \label{fig:appendix_image_results_dmc}
\end{figure}

\begin{figure}[t!]
    \centering
    \includegraphics[width=\textwidth]{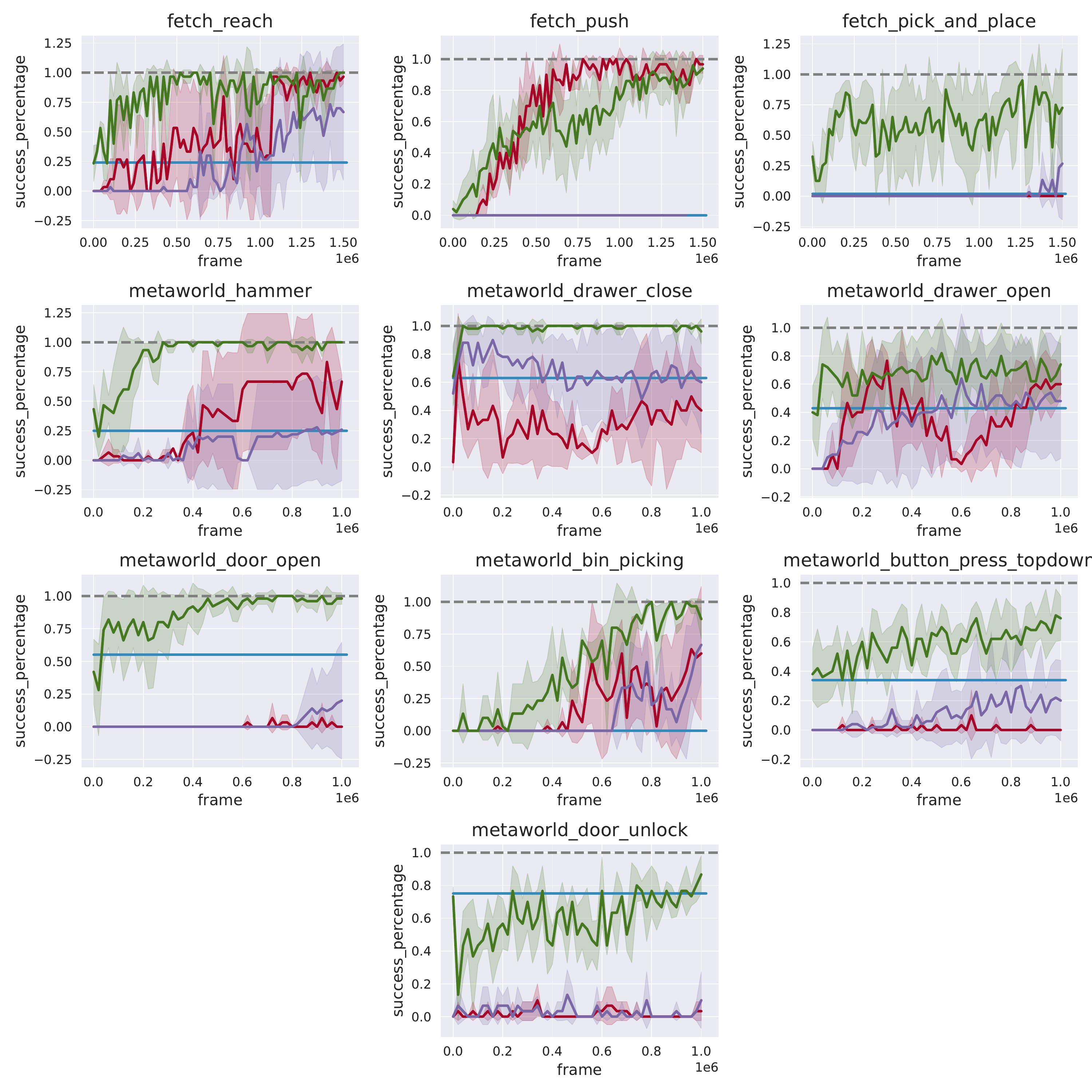}
    \cblock{128}{128}{128}\hspace{1mm}Expert\hspace{1.5mm}
    \cblock{52}{138}{189}\hspace{1mm}BC\hspace{1.5mm}
    \cblock{122}{104}{166}\hspace{1mm}OT\hspace{1.5mm}
    \cblock{166}{6}{40}\hspace{1mm}DAC\hspace{1.5mm}
    \cblock{70}{120}{33}\hspace{1mm}ROT (Ours)\hspace{1.5mm}
    \caption{Pixel-based continuous control learning on 3 OpenAI Gym Robotics and 7 Meta-World tasks. Shaded region represents $\pm1$ standard deviation across 5 seeds. We notice that ROT is significantly more sample efficient compared to prior work.}
    \label{fig:appendix_image_results_fetch_metaworld}
\end{figure}

\begin{figure}[t!]
    \centering
    \includegraphics[width=\textwidth]{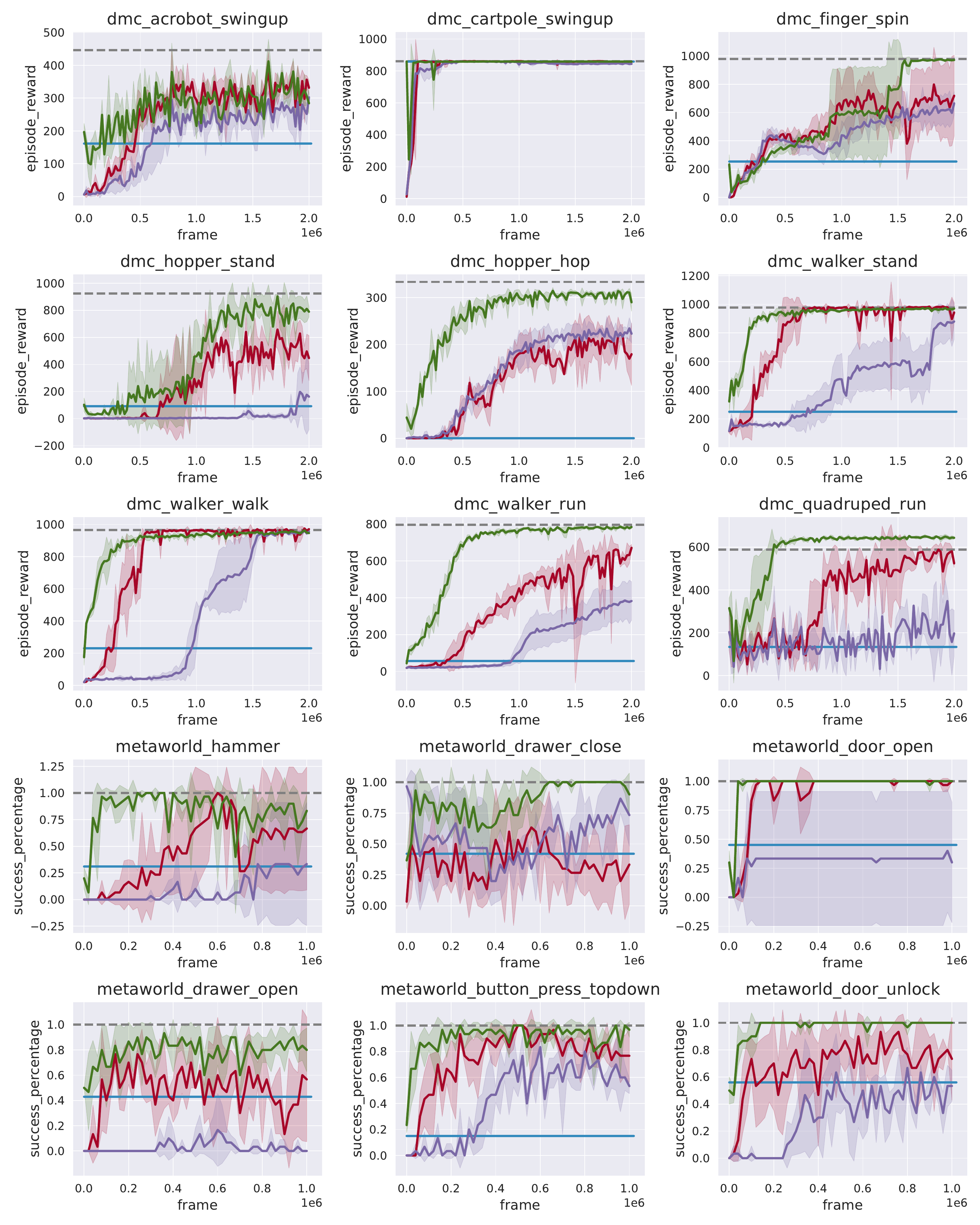}
    \cblock{128}{128}{128}\hspace{1mm}Expert\hspace{1.5mm}
    \cblock{52}{138}{189}\hspace{1mm}BC\hspace{1.5mm}
    \cblock{122}{104}{166}\hspace{1mm}OT\hspace{1.5mm}
    \cblock{166}{6}{40}\hspace{1mm}DAC\hspace{1.5mm}
    \cblock{70}{120}{33}\hspace{1mm}ROT (Ours)\hspace{1.5mm}
    \caption{State-based continuous control learning on DMC and Meta-World tasks. We notice that ROT is significantly more sample efficient compared to prior work.}
    \label{fig:appendix_state_results}
\end{figure}

In addition to the results provided in Sec.~\ref{sec:eff}, Fig.~\ref{fig:appendix_image_results_dmc} and Fig.~\ref{fig:appendix_image_results_fetch_metaworld} shows the performance of ROT for pixel-based imitation on 10 tasks from the DeepMind Control suite, 3 tasks from the OpenAI Robotics suite and 7 tasks from the Meta-world suite. On all but one task, ROT is significantly more sample efficient than prior work. Finally, the improvements from ROT hold on state-based observations as well(see Fig.~\ref{fig:appendix_state_results}). Table~\ref{tab:speedup_factor} provides a comparison between the factor of speedup of ROT to reach 90\% of expert performance compared to prior state-of-the-art~\cite{kostrikov2018discriminator,cohen2022imitation} methods.

\begin{table}[h!]
    \begin{center}
    \setlength{\tabcolsep}{5pt}
    \renewcommand{\arraystretch}{1.5}
    \begin{tabular}{ c c c c c} 
     \hline
     Suite & Tasks & ROT & 2nd Best Model & Speedup Factor \\
     \hline
     DeepMind Control   & Acrobot Swingup  & 200k & 600k (OT) & 3\\ 
                        & Cartpole Swingup & 100k & 350k (OT) & 3.5\\
                        & Finger Spin      & 20k  & 700k (OT) & 35\\
                        & Cheetah Run      & 400k & 2M   (DAC)& 5\\
                        & Hopper Stand     & 60k. & 750k (OT) & 12.5\\
                        & Hopper Hop       & 200k & $>$2M   (DAC)& 10\\
                        & Walker Stand     & 80k  & 400k (DAC)& 5\\
                        & Walker Walk      & 200k & 750k (DAC)& 3.75\\
                        & Walker Run       & 320k & $>$2M   (OT)& 6.25\\
                        & Quadruped Run    & 400k & $>$2M   (DAC) & 5\\
     \hline
     OpenAI Robotics & Fetch Reach          & 300k & 1.1M (DAC)& 3.67\\ 
                     & Fetch Push           & 1.1M & 600k (DAC)& 0.54\\
                     & Fetch Pick and Place & 750k & $>$1.5M (OT) & 2\\
     \hline
     Meta-World      & Hammer                 & 200k & $>$1M (DAC)& 5\\ 
                     & Drawer Close           & 20k  & $>$1M (OT) & 50\\
                     & Drawer Open            & $>$1M & $>$1M (OT) & 1\\
                     & Door Open              & 400k & $>$1M (OT) & 2.5\\
                     & Bin Picking            & 700k & $>$1M (OT) & 1.43\\
                     & Button Press Topdown   & $>$1M & $>$1M (OT) & 1\\
                     & Door Unlock            & 1M & $>$1M (OT) & 1\\
    \hline
    \end{tabular}
    \end{center}
    \caption{Task-wise comparison between environment steps required to reach 90\% of expert performance for pixel-based ROT compared to the strongest baseline for each task.}
    \label{tab:speedup_factor}
\end{table}

\subsection{Does soft Q-filtering improve imitation?}
\label{sec:appendix_sqf-exp}
\begin{figure}[t!]
    \centering
    \includegraphics[width=\textwidth]{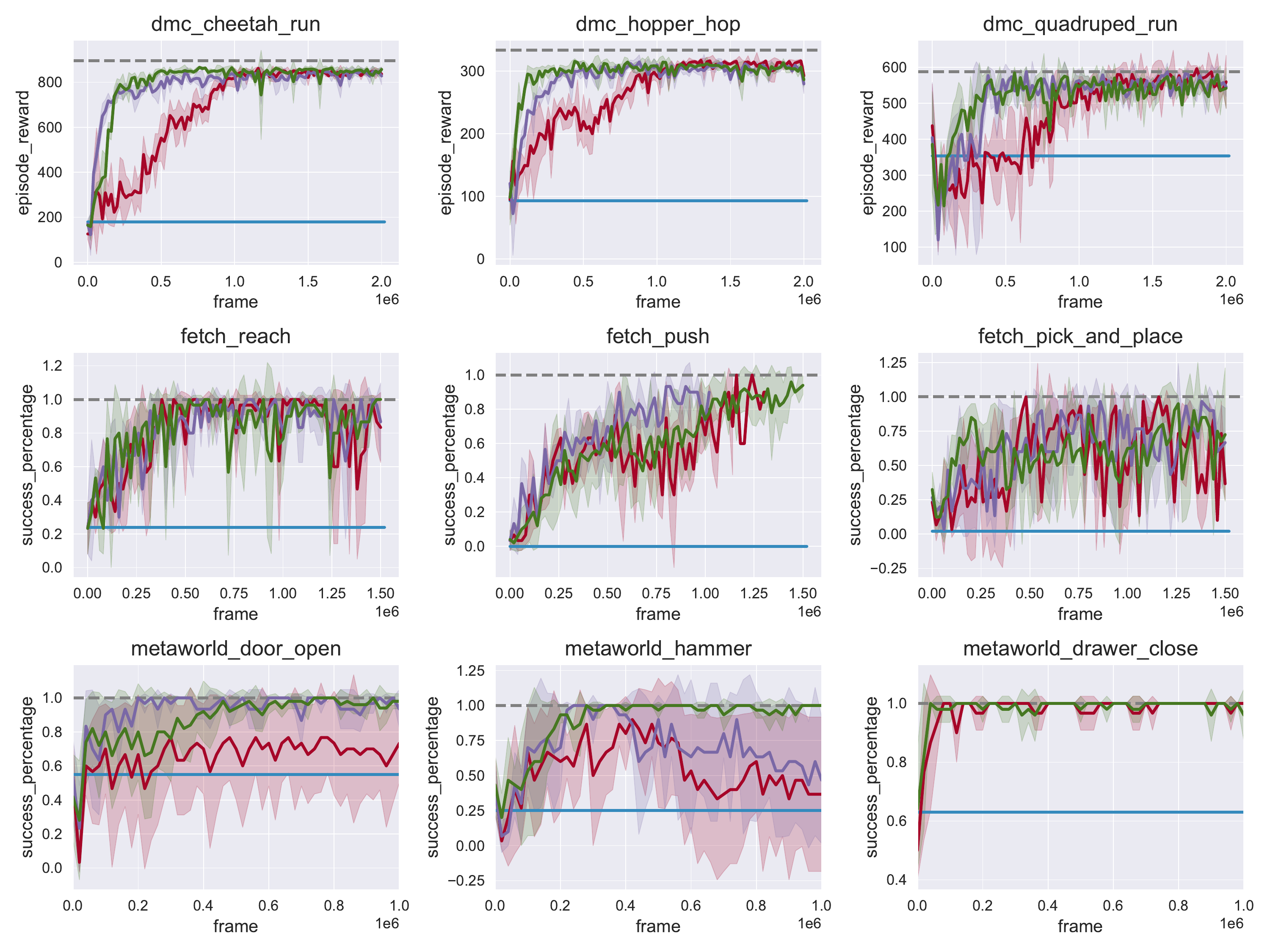}
    \cblock{128}{128}{128}\hspace{1mm}Expert\hspace{1.5mm}
    \cblock{52}{138}{189}\hspace{1mm}BC\hspace{1.5mm}
    \cblock{166}{6}{40}\hspace{1mm}Finetune with fixed weight\hspace{1.5mm}
    \cblock{122}{104}{166}\hspace{1mm}Finetune with fixed schedule\hspace{1.5mm}
    \cblock{70}{120}{33}\hspace{1mm}ROT (Ours)\hspace{1.5mm}
    \caption{Pixel-based ablation analysis on the effect of varying BC regularization schemes. We observe that our adaptive soft-Q filtering regularization is more stable compared to prior hand-tuned regularization schemes.}
    \label{fig:appendix_importance_of_sqf}
\end{figure}

Extending the results shown in Fig.~\ref{fig:importance_of_sqf}, we provide training curves from representative tasks in each suite in Fig.~\ref{fig:appendix_importance_of_sqf}. We observe that our adaptive soft-Q filtering regularization is more stable compared to prior hand-tuned regularization schemes. ROT is on par and in some cases exceeds the efficiency of a hand-tuned decay schedule, while not having to hand-tune its regularization weights. 

\subsection{How does ROT compare to standard reward-based RL?}
\label{appendix:reward_rl}
\begin{figure}[t!]
    \centering
    \includegraphics[width=\textwidth]{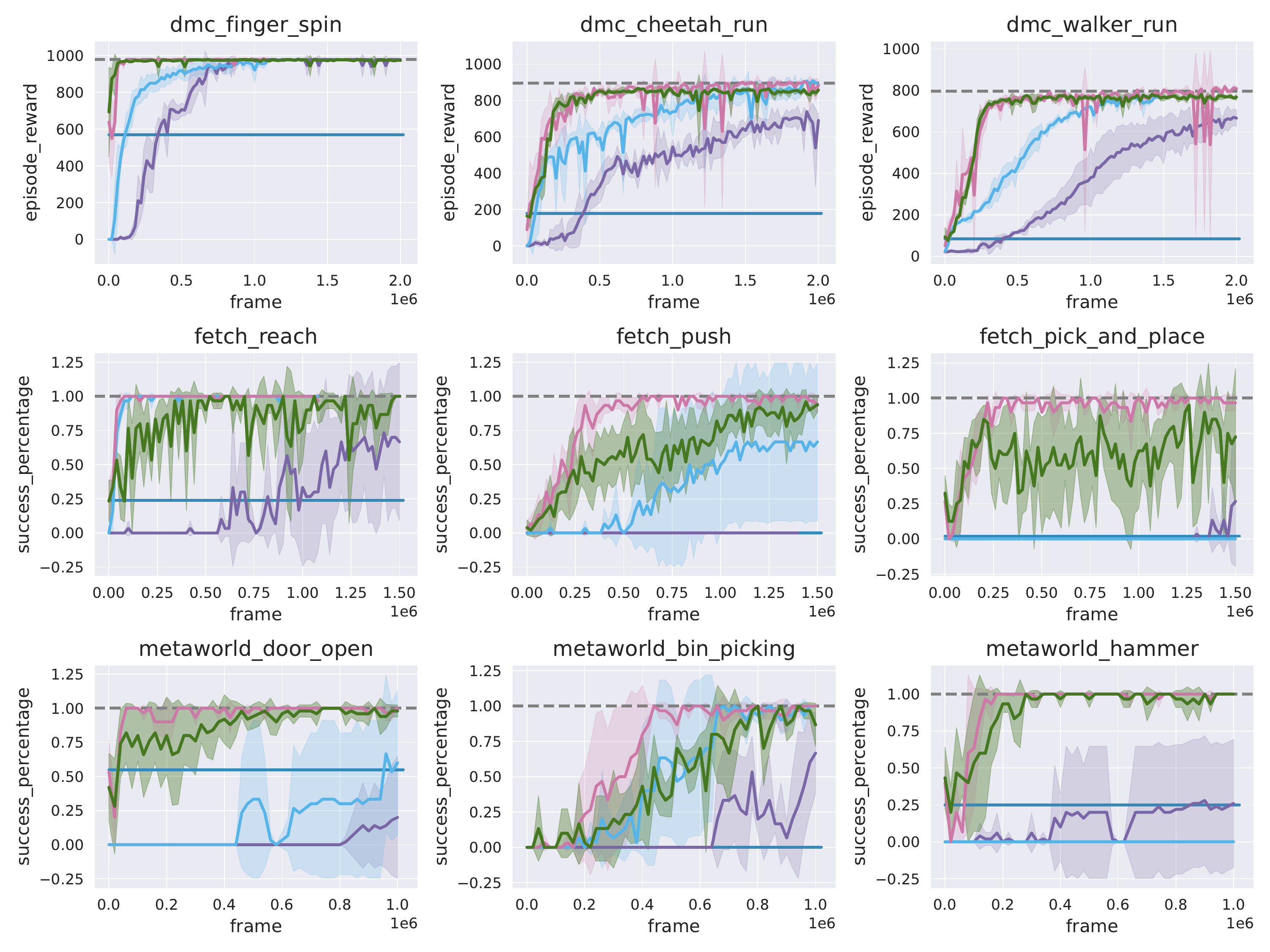}
    \cblock{128}{128}{128}\hspace{1mm}Expert\hspace{1.5mm}
    \cblock{52}{138}{189}\hspace{1mm}BC\hspace{1.5mm}
    \cblock{122}{104}{166}\hspace{1mm}OT\hspace{1.5mm}
    \cblock{86}{180}{233}\hspace{1mm}DrQ-v2(RL)\hspace{1.5mm}
    \cblock{204}{121}{167}\hspace{1mm}Demo-DrQ-v2\hspace{1.5mm}
    \cblock{70}{120}{33}\hspace{1mm}ROT (Ours)\hspace{1.5mm}
    \caption{Pixel-based ablation analysis on the performance comparison of ROT against DrQ-v2, a reward-based RL method. Here we see that ROT can outperform plain RL that requires explicit task-reward. However, we also observe that this RL  method combined with our regularization scheme provides strong results.}
    \label{fig:appendix_reward_rl}
\end{figure}

Extending the results shown in Fig.~\ref{fig:choice_of_irl_reward_rl}, we provide training curves from representative tasks in each suite in Fig.~\ref{fig:appendix_reward_rl}, thus showing that ROT can outperform standard RL that requires explicit task-reward. We also show that this RL method combined with our regularization scheme (represented by Demo-DrQ-v2 in Fig.~\ref{fig:appendix_reward_rl} provides strong results.

\subsection{How important are the design choices in ROT?}
\label{appendix:ablations}

\begin{figure}[t!]
    \centering
    \includegraphics[width=.98\linewidth]{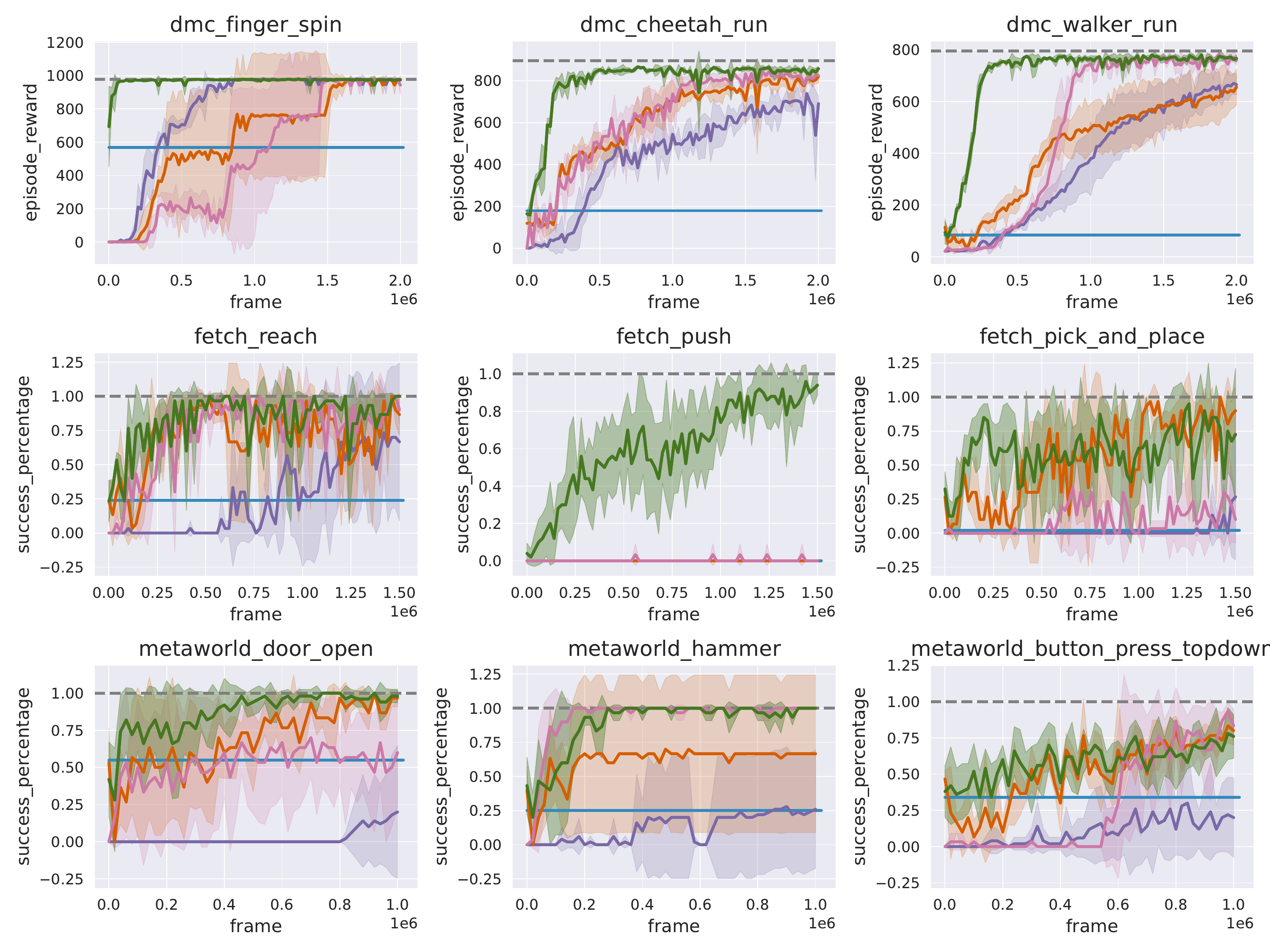}
    \cblock{52}{138}{189}\hspace{1mm}BC\hspace{1.5mm}
    \cblock{122}{104}{166}\hspace{1mm}OT\hspace{1.5mm}
    \cblock{213}{94}{0}\hspace{1mm}BC+OT\hspace{1.5mm}
    \cblock{204}{121}{167}\hspace{1mm}OT+BC Reg.\hspace{1.5mm}
    \cblock{70}{120}{33}\hspace{1mm}ROT (Ours)\hspace{1.5mm}
    \caption{Pixel-based ablation analysis on the importance of pretraining and regularizing the IRL policy. The key takeaway from these experiments is that both pretraining and BC regularization are required to obtain sample-efficient imitation learning.}
    \label{fig:appendix_ablation_pretraining}
\end{figure}

\paragraph{Importance of pretraining and regularizing the IRL policy} Fig.~\ref{fig:appendix_ablation_pretraining} compares the following variants of ROT on set of pixel-based tasks: (a) Training the IRL policy from scratch (OT); (b) Finetuning a pretrained BC policy without BC regularization (BC+OT); (c) Training the IRL policy from scratch with BC regularization (OT+BC Reg.). We  observe that pretraining the IRL policy (BC+OT) does not provide a significant difference without regularization. This can be attributed to the `forgetting behavior' of pre-trained policies, studied in ~\citet{nair2020awac}. Interestingly, we see that even without BC pretraining, keeping the policy close to a behavior distribution (OT+BC Reg.) can yield improvements in efficiency over vanilla training from scratch. Our key takeaway from these experiments is that both pretraining and BC regularization are required to obtain sample-efficient imitation learning.

\begin{figure}[t!]
    \centering
    \includegraphics[width=.98\linewidth]{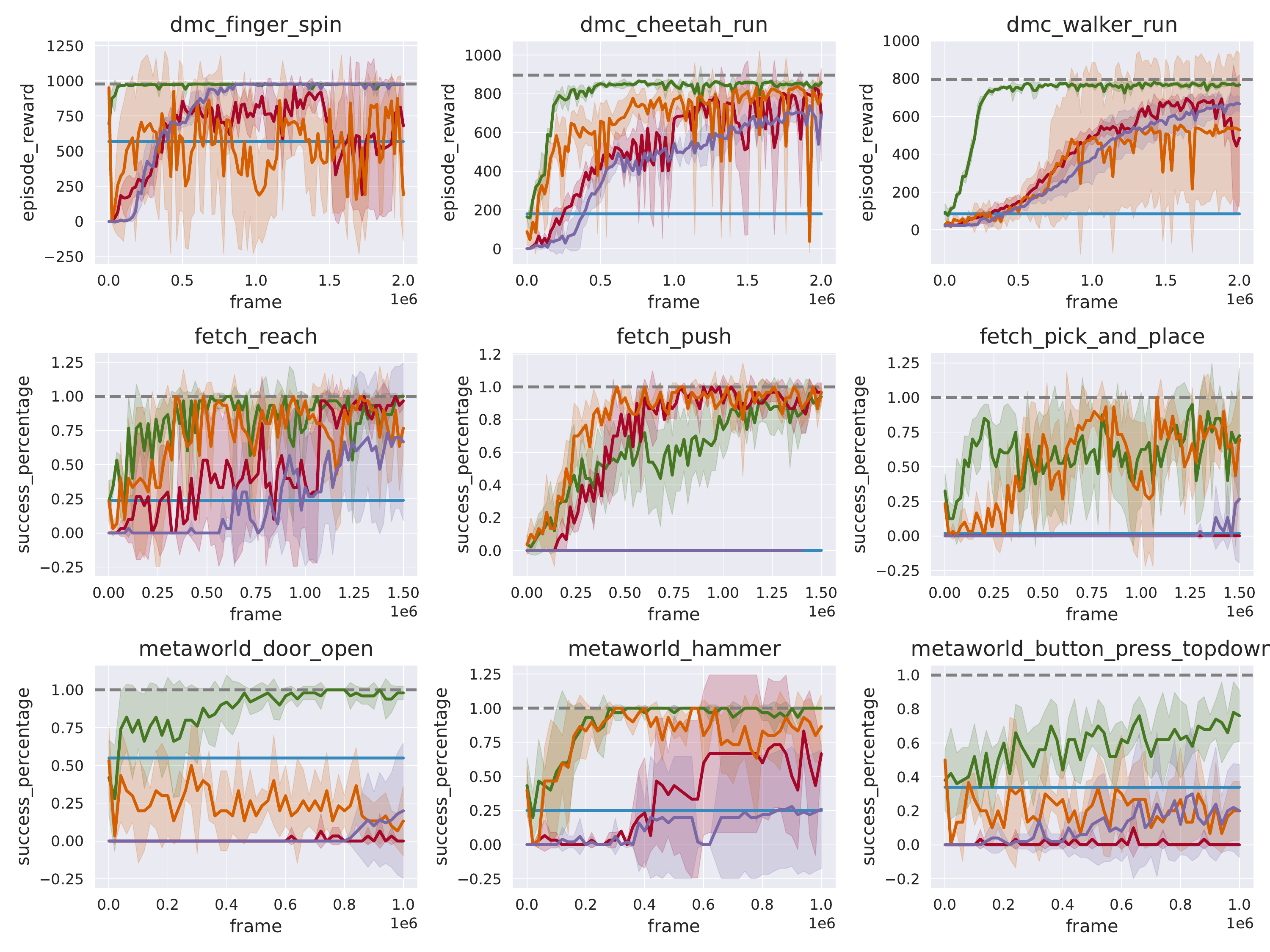}
    \cblock{128}{128}{128}\hspace{1mm}Expert\hspace{1.5mm}
    \cblock{52}{138}{189}\hspace{1mm}BC\hspace{1.5mm}
    \cblock{166}{6}{40}\hspace{1mm}DAC\hspace{1.5mm}
    \cblock{122}{104}{166}\hspace{1mm}OT\hspace{1.5mm}
    \cblock{213}{94}{0}\hspace{1mm}RDAC\hspace{1.5mm}
    \cblock{70}{120}{33}\hspace{1mm}ROT (Ours)\hspace{1.5mm}
    \caption{Pixel-based ablation analysis on the choice of base IRL method. We find that although adversarial methods benefit from regularized BC, the gains seen are smaller compared to ROT.}
    \label{fig:appendix_choice_of_irl}
\end{figure}

\paragraph{Choice of IRL method} In ROT, we build on OT-based IRL instead of adversarial IRL. This is because adversarial IRL methods require iterative reward learning, which produces a highly non-stationary reward function for policy optimization. In  Fig.~\ref{fig:appendix_choice_of_irl}, we compare ROT with adversarial IRL methods that use our pretraining and adaptive BC regularization technique (RDAC). We find that our soft Q-filtering method does improve prior state-of-the-art adversarial IRL (RDAC vs. DAC in Fig.~\ref{fig:appendix_choice_of_irl}). However, our OT-based approach (ROT) is more stable and on average leads to more efficient learning.

\end{document}